\documentclass[pdflatex,sn-mathphys-ay]{sn-jnl}

\usepackage[utf8]{inputenc}
\usepackage[UKenglish]{babel}

\usepackage{amsmath,amssymb,amsfonts}
\usepackage{graphicx}
\usepackage{textcomp}
\usepackage{xcolor}
\usepackage{xspace}

\usepackage{manyfoot}%
\usepackage{listings}%
\usepackage{mathrsfs}%
\usepackage[title]{appendix}%
\usepackage{amsthm}
\usepackage{booktabs}
\usepackage[separate-uncertainty=true,retain-zero-uncertainty=true]{siunitx}
\usepackage[plain]{algorithm}
\usepackage[noend]{algpseudocode}
\usepackage[nameinlink, noabbrev, capitalise]{cleveref}
\usepackage{tikz}
\usetikzlibrary{arrows.meta}

\let\orcidlogo\undefined 
\usepackage{orcidlink}

\definecolor{oxblue}{cmyk}{1, 0.87, 0.42, 0.51}
\definecolor{oxlilac}{cmyk}{0.19, 0.22, 0, 0}
\definecolor{oxgreen}{cmyk}{0.54, 0, 1, 0}
\definecolor{oxpink}{cmyk}{0.02, 0.58, 0.51, 0}
\definecolor{oxred}{cmyk}{0.18, 1.0, 0.74, 0.08}
\definecolor{oxaqua}{cmyk}{0.84, 0, 0.33, 0}
\definecolor{oxyellow}{cmyk}{0.08, 0, 0.69, 0}

\AtBeginEnvironment{algorithmic}{\small\hrule\vspace{2mm}}
\AtEndEnvironment{algorithmic}{\vspace{2mm}\hrule}
\algrenewcommand{\algorithmiccomment}[1]{\hfill\makebox[4.4cm][l]{\color{gray!80!orange}$\rangle\rangle$ #1}}
\algrenewcommand{\alglinenumber}[1]{\color{gray!80}\footnotesize#1:}

\newcommand{\hl}[1]{{\setlength{\fboxsep}{0pt}\colorbox{oxgreen!25}{\strut #1}}}
\algrenewcommand\algorithmicindent{1.2em}

\makeatletter
\AddToHook{env/algorithmic/before}{\def\@currentcounter{ALG@line}}
\makeatother
\crefname{line}{line}{lines}
\crefalias{ALG@line}{line}

\newcommand{\bm}{\fontseries{b}\selectfont}

\newcommand{\inputdomain}{\mathbb{I}}
\newcommand{\outputdomain}{\mathbb{O}}
\newcommand{\polytopeset}{\mathbb{D}}
\newcommand{\polytope}{\mathbf{P}}
\newcommand{\subdomain}{\inputdomain_\text{sub}}
\newcommand{\samples}{\mathbf{X}}
\newcommand{\xv}{\mathbf{x}}
\newcommand{\lx}{\underline{\mathbf{x}}}
\newcommand{\ux}{\overline{\mathbf{x}}}
\newcommand{\zv}{\mathbf{z}}
\newcommand{\lz}{\underline{\mathbf{z}}}
\newcommand{\uz}{\overline{\mathbf{z}}}
\newcommand{\lA}{\underline{\mathbf{A}}}
\newcommand{\lb}{\underline{\mathbf{b}}}
\newcommand{\uA}{\overline{\mathbf{A}}}
\newcommand{\ub}{\overline{\mathbf{b}}}
\newcommand{\betacrown}{\mbox{$\beta$-CROWN}\xspace}
\DeclareMathOperator*{\argmax}{arg\,max}

\def\dottedbox{\tikz[baseline]\node[draw=black,dash pattern=on 2pt off 2pt,inner sep=5pt,yshift=3pt] {\vphantom{\enspace}};}

\begin{document}

\title{Efficient Preimage Approximation for Neural Network Certification}

\author*[1]{\fnm{Anton}~\sur{Björklund} \orcidlink{0000-0002-7749-2918}}\email{anton.bjorklund@cs.ox.ac.uk}
\author[2]{\fnm{Mykola}~\sur{Zaitsev} \orcidlink{0000-0002-7547-5284}}
\author[3]{\fnm{Paolo}~\sur{Morettin} \orcidlink{0000-0003-4321-5215}}
\author[1]{\fnm{Marta}~\sur{Kwiatkowska} \orcidlink{0000-0001-9022-7599}}
\affil[1]{\orgdiv{Department of Computer Science}, \orgname{University of Oxford}, \country{United Kingdom}}
\affil[2]{\orgdiv{Department of Computer Science}, \orgname{Johannes Kepler University Linz}, \country{Austria}}
\affil[3]{\orgdiv{Department of Information Engineering and Computer Science}, \orgname{University of Trento}, \country{Italy}}

\abstract{
    The growing reliance on artificial intelligence in safety- and security-critical applications is
    raising concerns about the robustness of neural networks to erroneous or adversarial input.
    Certification is a methodology for ensuring model trustworthiness by providing formal guarantees on model behaviour.
    While most verification methods focus on worst-case analysis by bounding the network output, an alternative approach based on approximating the preimage can complement such analysis
    by estimating the proportion of inputs that satisfy a given specification.
    However, existing preimage-based methods, such as the state-of-the-art PREMAP, are 
    limited to fully connected neural networks of moderate dimensionality.
    In this paper, we introduce PREMAP2, a collection of
    algorithmic extensions to PREMAP that enhance its scalability and efficiency through improved branching heuristics, adaptive Monte Carlo sampling, and reverse bound propagation.
    We further endow PREMAP2 with additional functionality such as support for non-uniform priors and confidence intervals.
    These advances enable the application of PREMAP2 to previously intractable settings, including real-world patch attacks against convolutional neural networks, 
    where adversarial stickers or lighting conditions obscure parts of images.
    We showcase the effectiveness of our approach across several use cases, including certifying reliability, robustness, interpretability, and fairness, on domains  ranging from computer vision  to control tasks.
    Our implementation is available as open-source software.
}


\maketitle

\section{Introduction} \label{sec:intro}

\begin{figure}[htb]
    \centering
    \includegraphics[width=12cm]{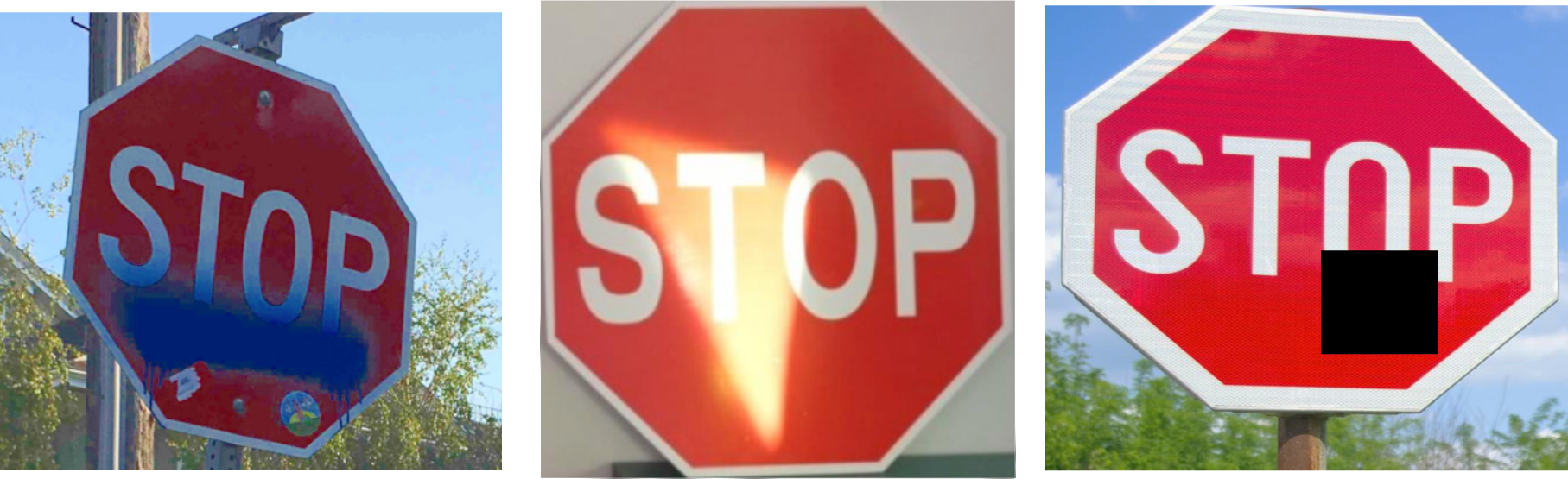}
    \caption{Examples of physical patch attacks. Left: real graffiti \citep{eykholt2018robustphysicalworldattacksdeep}. Middle: sunbeam \citep{cao2024secure}. Right: our abstraction of a patch attack.}
    \label{fig:ex1}
\end{figure}

The widespread deployment of AI technologies places significant demands on their trustworthiness and reliability, which are of particular concern in high-stakes applications, as recognized in the EU AI Act~\citep{eu2024ai}.
Much effort has been devoted to ensuring the safety and security of AI applications involving neural networks (NNs) due to their instability to adversarial perturbations \citep{goodfellow2014explaining}.
A particularly challenging real-world example are ``patch attacks'' on images, such as traffic signs. 
In Figure \ref{fig:ex1}, we show two examples of physical patch attacks: real graffiti on a traffic sign \citep[left,][]{eykholt2018robustphysicalworldattacksdeep} and a patch caused by a sunbeam \citep[middle,][]{cao2024secure}. 

The examples above highlight the importance of studying the robustness of 
models to erroneous and adversarial inputs for applications such as safe autonomous driving or biometric security software.
To this end, a wide range of methodologies have been developed for neural networks to enhance their robustness, either via adversarial training \citep{shi2021fast,mao2024understanding} or designing certifiable \citep{xiang2024patchcure} or empirical \citep{wei2024physical} defences.
An alternative method is to certify robustness by computing deterministic \citep{huang2017safety,katz2017reluplex} or probabilistic \citep{webb2018statistical,yang2021quantitative,cohen2019certified} guarantees on the network outputs.
Similarly to formal verification, certification is more powerful than adversarial training or empirical defences because it can guarantee that the network output is correct for all inputs in a given region.
Using the patch attack in Figure \ref{fig:ex1} (right) as an illustration, 
certification aims to guarantee the prediction
in the presence of any patch pattern in the black square, that is, formed using all possible colour combinations for the pixels.

Many robustness certification methods bound the worst-case output of the neural network when proving robustness for a given input domain.
An alternative approach, which gives more granular information, focuses on bounding the preimage of a given output specification (typically a polyhedron), either through over-approximating the preimage \citep{kotha2023provably,zhang2024provable,zhang2024premap,boetius2025solving} or under-approximating it \citep{zhang2024provable,zhang2024premap,boetius2025solving}.
Under-approximation offers stronger compliance guarantees than over-approximations, since all inputs in the computed approximation satisfy the guarantee.
Under-approximations also provide quantitative verification, i.e., estimating the proportion of inputs that meet the specification.

Several preimage approximation methods \citep{kotha2023provably,boetius2025solving} rely on dividing the input domain to obtain tighter approximations, which makes higher-dimensional inputs, such as images, intractable.
The state-of-the-art algorithm PREMAP \citep{zhang2024provable,zhang2024premap} offers the option to split the domain on non-linear activations, as they are the source of looser bounds \citep{xu2020automatic}.
This strategy works better for higher-dimensional inputs;  
however, PREMAP still struggles with convolutional neural networks (CNN) as they have significantly more non-linear activations compared to fully connected (FC) networks. 

In this paper, we present PREMAP2, an improvement of the
PREMAP algorithm aimed at broadening its applicability to a wider range of use cases.
To this end, we improve the computational performance of PREMAP
by significantly reducing the size of the computation tree
through calculating tighter bounds, carefully selecting which neuron to branch on, and avoiding heavier optimizations when possible.
In particular, PREMAP2 enables
preimage approximations of convolutional neural networks, which could not be handled previously, and provides new capabilities in the form of non-uniform priors and confidence intervals for the preimage volumes.

We demonstrate through empirical comparison that PREMAP2 consistently outperforms the original PREMAP \citep{zhang2024premap} on both under- and over-approximations, achieving
running time improvements of at least an order of magnitude on reinforcement learning control tasks, while on patch attack certification the time-limited completion rates increase threefold for non-trivial cases.
We also showcase PREMAP2 on a range of case studies: (i) certification for patch attacks
on convolutional networks;
(ii) out-of-distribution (OOD) input detection, where we show the utility of both under- and over-approximations to certify
outliers; (iii) explainability (XAI), where we demonstrate how PREMAP can complement existing explainability techniques in analysing and debugging neural network models by highlighting important regions in the input;
and (iv) fairness, where we compare outcomes for different groups with a prior.    
Our results demonstrate the potential of preimage approximation methodology for exploitation in reliability and robustness analysis and certification.

\section{Related works} \label{sec:related}

Robustness certification methods focus on computing deterministic \citep{huang2017safety,katz2017reluplex} or probabilistic/statistical \citep{webb2018statistical,yang2021quantitative} guarantees on the output of a neural network.
We can classify the methods into complete methods, such as constraint solving \citep{katz2017reluplex}, or sound but incomplete methods, e.g., convex relaxation \citep{salman2019convex}, which can be strengthened to completeness by employing branch-and-bound procedures \citep{bunel2018unified,Bunel20BaB}. 

Building on \citet{zhang2024provable, zhang2024premap}, our work employs convex (linear) relaxation and Monte Carlo sampling in a branch-and-refine framework for approximating the preimage of the neural network with exact (not statistical) quantitative guarantees. We leverage LiRPA \citep{xu2020automatic} as implemented in CROWN tools \citep{zhang2018efficient,xu2021fast,wang2021betacrown} to compute symbolic relaxation bounds, adapting to preimage approximation.

We note that exact preimage computation is intractable for high dimensions, although a variant of this problem known as backward reachability has been studied in control \citep{vincent2021reachable} through exact computation or (guaranteed) over-approximation \citep{rober2023backward,Dathathri19Inverse}.
Compared to \citet{bunel2018unified,Bunel20BaB}, our branch-and-refine approach focuses on minimizing the volume difference between the approximation and the preimage rather than maximizing a function value. 
The methods of \citet{kotha2023provably,boetius2025solving} also perform preimage approximations, but the former only over-approximations. 
However, they both \citep{kotha2023provably,boetius2025solving} use input-space splitting, which scales poorly beyond low-dimensional data.

Methods defending against adversarial attacks \citep{wei2024physical,xiang_short_2023} include active defences, such as small receptive fields or masking out all adversarial pixels from the input image, and passive defences, for example, certifiable training using bound propagation. 
Small receptive fields with robust aggregation provide
improved robustness at the cost of worse clean performance, where \citet{xiang2024patchcure} offers a computationally efficient trade-off.
Certifiable training \citep{mao2024understanding} involves estimating and propagating the bounds of activations in each layer to bound the attacker’s influence on final predictions and is computationally expensive.
Randomized smoothing \citep{cohen2019certified} consists of repeated predictions with random noise added to the input during inference or training \citep{fang2025multihead}, which yields 
probabilistic certification of robustness.
In contrast, PREMAP2 provides quantitative certification of robustness by approximating the set of inputs for which the prediction is \emph{guaranteed} to be robust for already trained networks.

\section{Preliminaries} \label{sec:prelim}

In this section, we define preliminary concepts concerning neural networks,   including linear relaxation and preimage under- and over-approximation.  

\subsection{Neural networks}

We use $f : \mathbb{R}^d \rightarrow \mathbb{R}^m$ to denote a feed-forward neural network, where $f(\mathbf{x}) = \mathbf{y}$.
For layer $l$, $\mathbf{W}^l$ denotes the weight matrix, $\mathbf{b}^l$ the bias, $\mathbf{z}^l$ the pre-activation neurons, and $h^l(\mathbf{z}^l)$ the activation function, such that $\mathbf{z}^l = \mathbf{W}^l h^{l-1}(\mathbf{z}^{l-1}) + \mathbf{b}^l$.
We use under- and over-lines to denote the lower and upper bounds, e.g., $\lx_i \leq \xv_i \leq \ux_i$.
In this paper we focus on ReLU neural networks, where $ h^l(\mathbf{z}^l) = \text{ReLU}(\mathbf{z}^l) = \max(\mathbf{z}^l, 0) $ is applied element-wise. However, our method can also be generalized to other non-linear activation functions that can be bounded by linear functions \citep{zhang2018efficient,chevalier2024achieving}.

\subsection{Linear relaxation}

Linear relaxation is used to transform non-convex neurons into linear programs, which provide effective means to approximate a neural network to ensure tractability of analysis. As shown in \cref{fig:relu}, given concrete lower and upper bounds $\lz^l \leq \mathbf{z}^l \leq \uz^l$ on the pre-activation values of layer $l$, there are three cases to consider. 
When $\uz^l_i \leq 0$ then $h^l(\mathbf{z}^l_i) = 0$ and the neuron is \textit{inactive}.
Similarly, when  $\lz^l_i \geq 0$ then $h^l(\mathbf{z}^l_i) = \mathbf{z}^l_i$ and the neuron is \textit{active}.
Otherwise, the neuron is \textit{unstable} and can be bounded by
\begin{equation}
\label{eq:relu_bounds}    
\alpha \mathbf{z}^l_i \leq h^l(\mathbf{z}^l_i) \leq (\mathbf{z}^l_i - \lz^l_i) {\uz^l_i}/{(\uz^l_i-\lz^l_i)},
\end{equation}
where $\alpha$ is any value in $[0, 1]$.

\begin{figure}[htb]
    \newcommand{\figaxis}[3]{
        \draw (0, -0.65) node {#1};
        \draw[black!60, ->] (#2,0) -- (#3,0);
        \draw[black!60, <-] (0,1) -- (0,-0.5);
        \draw[very thick] (#2,0) -- (0,0) -- (#3-0.1,#3-0.1);
        \draw (#3, 0.2) node {$\mathbf{z}^l_i$};
        \draw (0.1, 1.2) node {$h^l(\mathbf{z}^l_i)$};
    }
    \newcommand{\figlimits}[3][]{
        \draw[dashed, oxred] (#2,-0.15) -- (#2,1);
        \draw[dashed, oxred] (#3,-0.15) -- (#3,1);
        \draw[oxred] (#2+0.02#1, -0.3) node {$\lz^l_i$};
        \draw[oxred] (#3+0.05, -0.32) node {$\uz^l_i$};
    }
    \centering
    \begin{tikzpicture}[every node/.style={scale=0.9}]
    \tikzset{scale=1.3}
        \figaxis{Inactive}{-1.1}{0.7}
        \draw[ultra thick, oxred] (-0.9,0) -- (-0.3,0);
        \figlimits{-0.9}{-0.3}
        
        \tikzset{shift={(1.7,0)}}
        \figaxis{Active}{-0.6}{1.0}
        \draw[ultra thick, oxred] (0.2,0.2) -- (0.7,0.7);
        \figlimits{0.2}{0.7}
        
        \tikzset{shift={(2.4,0)}}
        \path[fill=oxpink] (-0.5,0) -- (0.5,0.5) -- (0.5,0.15) -- (-0.5,-0.2) -- (-0.5,0);
        \figaxis{Active}{-0.8}{0.8}
        \draw[ultra thick, oxred] (-0.5,0) -- (0.5,0.5);
        \draw[ultra thick, oxred] (-0.5,-0.18) -- (0.5,0.15);
        \figlimits[-0.16]{-0.5}{0.5}
    \end{tikzpicture}
    \caption{Linear bounds for inactive, active and unstable ReLU neurons.}
    \label{fig:relu}
\end{figure}

By relaxing all non-linear activations,
we can
compute linear lower and upper bounds for the whole neural network $f$:
\begin{equation}
    \label{eq:lirpa}
    \lA \xv + \lb \leq f(\xv) \leq \uA \xv + \ub,
\end{equation}
for a given input region.
These methods are known as Linear Relaxation based Perturbation Analysis \citep[LIRPA,][]{xu2020automatic} algorithms.

\subsection{Preimage approximation}

Given a subset $ \outputdomain \subset \mathbb{R}^m $,
the \emph{preimage} of a function $ f : \mathbb{R}^d \to \mathbb{R}^m $ is defined to be the set of all inputs $ \mathbf{x} \in \mathbb{R}^d $ that are mapped by $f$ to an element of $\outputdomain$. Since computing the exact preimage is intractable, we are interested in computing an \emph{under-} or \emph{over-approximation} of the preimage.

To compute preimage approximations, we restrict the shape of the input and output domains to polyhedra and apply LiRPA.
We first 
bound the input domain $\inputdomain \subset \mathbb{R}^d$, for example as an axis-aligned hyper-rectangle $\inputdomain \subseteq \{ \mathbf{x} \in \mathbb{R}^d \mid \bigwedge_{i=1}^d \lx_i \leq \xv_i \leq \ux_i \}$.
We specify the output domain $\outputdomain$ as a set of linear inequalities (a \emph{polytope}) on the output $\bigwedge_{j=1}^{K} (\mathbf{c}_j^\top \mathbf{y} + d_j \geq 0)$.
Since the inequalities are linear, they can be implemented by adding a linear layer to the neural network $f_\outputdomain(\mathbf{x})_j = \mathbf{c}_j^\top f(\mathbf{x}) + d_j$.
Then, by bounding $f_\outputdomain$ with LiRPA, such that $\lA\mathbf{x}+\lb \leq f_\outputdomain(\mathbf{x}) \leq \uA\mathbf{x}+\ub$, we obtain an \emph{under-approximation} of the preimage as
\begin{equation*}
    \mathbf{x}\in\inputdomain \wedge \lA\mathbf{x}+\lb\geq0 \implies f(\mathbf{x}) \in \outputdomain
\end{equation*}
and an \emph{over-approximation} as
\begin{equation*}
    \mathbf{x}\in\inputdomain \wedge f(\mathbf{x}) \in \outputdomain \implies \uA\mathbf{x}+\ub\geq0.
\end{equation*}

\section{Overview of the PREMAP framework} \label{sec:overview}

The PREMAP algorithm \citep{zhang2024premap, zhang2024provable} 
computes both under- and over-approximation of the preimage for a  neural network $f$ with a given output specification $\outputdomain$ restricted to the input domain $\inputdomain$.
In this section we provide an overview of PREMAP in conjunction with our proposed improvement, referred to as PREMAP2, which is discussed further in \cref{sec:methods}.

\subsection{PREMAP workflow} \label{sec:workflow}

At the high level, 
PREMAP \citep{zhang2024provable, zhang2024premap} iteratively splits the input domain into subdomains and verifies which subdomains are inside or outside the preimage using linear bounds.
This procedure is visualized in \cref{fig:premap} for a simple 2D example with a non-linear decision boundary for the target preimage.

\begin{figure}[htb]
    \centering
    \newcommand{\figspace}[1]{
        \draw (0.8, -0.25) node {#1};
        \draw[black!60, ->] (-0.15,0) -- (1.75,0);
        \draw[black!60, ->] (0,-0.15) -- (0,1.75);
        \draw (1.7, 0.2) node {$x_1$};
        \draw (0.25, 1.67) node {$x_2$};
        \draw[very thick] (0,0) -- (1.5,0) -- (1.5,1.5) -- (0,1.5) -- (0,0);
        \draw[ultra thick, dotted, oxblue] plot [smooth] coordinates {(1.4,-0.1) (1.35,0.2) (0.8,0.8) (0.8,1.2) (1.0,1.4) (1.05,1.65)};
    }
    \begin{tikzpicture}[every node/.style={scale=0.9}]
    \tikzset{scale=1.3}
        \figspace{Setup}
        \draw[oxblue, thick, >->] (1.25,-0.23) -- (2.15,-0.23);
        
        \tikzset{shift={(2,0)}}
        \path[fill=oxyellow] (0,0) -- (1.19,0) -- (0.45,1.5) -- (0.0,1.5) -- (0,0);
        \figspace{Optimize}
        \draw[ultra thick, oxred] (1.19,0) -- (0.45,1.5);
        \draw[oxblue, thick, >->] (1.5,-0.23) -- (2.4,-0.23);
        
        \tikzset{shift={(2,0)}}
        \path[fill=oxyellow] (0,0) -- (1.19,0) -- (0.45,1.5) -- (0.0,1.5) -- (0,0);
        \figspace{Split}
        \draw[ultra thick, oxred] (1.19,0) -- (0.45,1.5);
        \draw[very thick] (0,0.75) -- (1.5,0.75);
        \draw[oxblue, thick, >->] (1.25,-0.23) -- (2.15,-0.23);
        \draw[oxblue, thick, -<] plot [smooth] coordinates {(0.35,-0.23) (0.15,-0.23) (0.25,-0.4) (1.675,-0.4) (3.6,-0.4) (3.7,-0.23) (3.5,-0.23)};
        
        \tikzset{shift={(2,0)}}
        \path[fill=oxyellow] (0,0) -- (1.38,0) -- (0.78,0.75) -- (0.65,0.75) -- (0.89,1.5) -- (0.0,1.5) -- (0,0);
        \figspace{Optimize}
        \draw[ultra thick, oxred] (1.38,0) -- (0.78,0.75);
        \draw[ultra thick, oxred] (0.65,0.75) -- (0.89,1.5);
        \draw[very thick] (0,0.75) -- (1.5,0.75);
    \end{tikzpicture}
    \caption{PREMAP workflow for preimage under-approximation. The black rectangles represent the input domain, the dotted blue line the decision boundary of a neural network, the black line splitting into smaller subdomains, the red lines the linear relaxation bounds obtained with LiRPA, and the yellow area the under-approximation of the preimage computed by PREMAP.}
    \label{fig:premap}
\end{figure}

Intuitively, the procedure alternates between bounding with linear relaxation and splitting of the input domain into disjoint subdomains.
In \cref{fig:premap}, the leftmost plot shows the input domain (black rectangle) with the decision boundary (dotted line). 
The second plot from the left demonstrates a preimage under-approximation (yellow shading), where the red line represents the (optimized) linear bound $\lA\mathbf{x}+\lb\geq0$.
In the third plot, the preimage is then split (black line) into two subdomains, and the final plot shows the optimized (larger) preimage under-approximation, where the red lines show linear bounds computed for the subdomains.

\subsection{The PREMAP/PREMAP2 algorithms} \label{sec:premap}

Given a neural network $f$ and output specification $\outputdomain$, PREMAP computes an under- or over-approximation of preimage of $\outputdomain$ restricted to the input domain $\inputdomain$. 
\cref{alg:main} shows the PREMAP2 algorithm introduced here, which follows the same outline as the original PREMAP, with our modifications highlighted with green shading, which are discussed further in \cref{sec:methods}.
We now describe the details of \cref{alg:main} and its additional functions (\cref{alg:functions}).

Since the initial linear bound (\cref{alg:main}, \cref{alg:line:preimage})
might be very loose \citep{wang2021betacrown}, PREMAP2 iteratively partitions a subdomain $\subdomain \subseteq \inputdomain$ into two disjoint subdomains $\subdomain^-$ and $\subdomain^+$ (\crefrange{alg:line:loop}{alg:line:union}), see also the third subplot of \cref{fig:premap}, which can be approximated independently in parallel (see \cref{alg:line:left,alg:line:right}) with a branch-and-refine approach.

We select the subdomain with the largest difference between the exact preimage and the approximation for splitting (\cref{alg:main}, \cref{alg:line:pop}).
To estimate the volumes we use Monte Carlo estimates (\cref{alg:line:sample0,alg:line:sample}). 
The split can be either on the input, which partitions the input into two subdomains along some feature, or based on the pre-activation value of an intermediate unstable neuron \citep{zhang2024premap}, called ReLU splitting.
As this paper deals with larger input domains, for which input splitting performs less well, we will focus on the latter (\cref{alg:line:select}).

A key advantage of ReLU splitting is that it allows us to replace unstable neuron bounds with precise bounds. 
For an unstable ReLU neuron $h^l(\zv^l_j) = \max(0, \zv^l_j)$, we use linear relaxation to bound the post-activation value as in \cref{eq:relu_bounds}.
When a split is introduced, the neuron becomes stable in each subdomain, and the exact linear functions $h^l(\zv^l_j) = \zv^l_j$ and $h^l(\zv^l_j) = 0$ can be used in place of its linear relaxation.
This typically tightens the approximation on each subdomain as the linear relaxation errors for this unstable neuron are removed and substituted with the exact symbolic function.

The bounds, $\lA \xv + \lb$, for the new under-approximations (or $\uA \xv + \ub$ for over-approximations) are calculated using \betacrown \citep{wang2021betacrown} and optimized via projected gradient descent (\cref{alg:functions}, \cref{alg:line:preimage}) to maximize the approximation \citep{zhang2024premap}, using the Monte Carlo samples.
The algorithm can be terminated anytime (with a valid approximation), but otherwise stops upon reaching a preset threshold for the approximation ratio (\cref{alg:main}, \cref{alg:line:sample}).
The union of the preimage approximations $\polytope$ for the disjoint subdomains $\subdomain$ constitutes the preimage approximation, which corresponds to the areas shaded yellow in the rightmost plot of \cref{fig:premap}.

\begin{algorithm}
\newcommand{\Sample}{\textsc{Sample}}
\newcommand{\Approx}{\textsc{PreimageApprox}}
\begin{algorithmic}[1]
\Function{Premap2}{$f, \inputdomain, \outputdomain, v_\text{th}, n$}
    \State $\samples \gets \Sample(\inputdomain, 5n)$ \Comment{Sample extra when it is easy} \label{alg:line:sample0}
    \State $\polytope \gets \Approx(f_\outputdomain, \inputdomain, \samples)$ \label{alg:line:preimage} \Comment{See \cref{alg:functions}}
    \State $\polytopeset \gets \{(\polytope, \inputdomain, \samples)\}$ \Comment{Approxim., domain, samples}
    \While{$\textsc{EstimateApproxRatio}(\polytopeset, f_\outputdomain) < v_\text{th}$} \label{alg:line:loop}
        \State $\polytope, \subdomain, \samples \gets \textsc{Pop}(\polytopeset, \argmax_j(\textsc{Priority}(\polytopeset_j, f_\outputdomain)))$ \label{alg:line:pop}
        \State \hl{$\samples \gets \Sample(\subdomain, n, \samples)$ } \label{alg:line:sample} \Comment{See \cref{sec:sampling}}
        \State \hl{$l, i \gets \textsc{SelectNeuron}(f_\outputdomain, \subdomain, \samples)$ } \label{alg:line:select} \Comment{See \cref{sec:heuristics}}
        \State \hl{$\subdomain \gets \textsc{CheckShortcut}( \subdomain, \samples)$} \label{alg:line:shortcut}\Comment{See \cref{sec:short}}
        \State $\subdomain^-,\samples^-, \subdomain^+,\samples^+ \gets \textsc{SplitNeuron}(\subdomain, \samples, l, i)$ \label{alg:line:split}
        \State \hl{$\textsc{PropagateBounds}(\subdomain^-, \subdomain^+, l, i)$} \label{alg:line:bounds} \Comment{See \cref{sec:tighten}}
        \State $\polytope^- \gets \Approx(f_\outputdomain, \subdomain^-,  \samples^-)$ \label{alg:line:left}
        \State $\polytope^+ \gets \Approx(f_\outputdomain, \subdomain^+,  \samples^+)$ \label{alg:line:right}
        \State $\polytopeset \gets \polytopeset \cup \{ (\polytope^-, \subdomain^-, \samples^-), ~ (\polytope^+, \subdomain^+, \samples^+ ) \}$ \label{alg:line:union}
    \EndWhile
    \State \Return $\{\polytope \mid (\polytope, \cdot, \cdot) \in \polytopeset\}$
\EndFunction
\end{algorithmic}
\floatname{algorithm}{\scriptsize\textbf{Algorithm}}
\caption{\scriptsize
    The \textsc{Premap2} algorithm.
    The inputs are the neural network $f$, the input domain $\inputdomain$, the output specifications $\outputdomain$, the early stopping threshold $v_\text{th}$, and the number of samples $n$.
    Some functions are detailed in \cref{alg:functions}.
    The lines with improvements compared to  \citet{zhang2024premap} 
    are highlighted with green shading and are discussed in \cref{sec:methods}.
}\label{alg:main}
\end{algorithm}

\begin{algorithm}
\newcommand{\LSE}{\textsc{LSE}}
\newcommand{\lirpa}{\textsc{Lirpa}}
\begin{algorithmic}[1]
\Function{PreimageApprox}{$f_\outputdomain, \subdomain, \samples$}
    \State $\lA^{(\boldsymbol{\alpha})}, \lb^{(\boldsymbol{\alpha})} \gets \lirpa(f_\outputdomain, \subdomain)$ \Comment{Calculate linear bounds}
    \State $\boldsymbol{\alpha} \gets \argmax\limits_{\boldsymbol{\alpha}} \sum\limits_{\mathbf{x} \in \samples} \sigma(-\log(\sum \exp({-\lA^{(\boldsymbol{\alpha})} \mathbf{x} - \lb^{(\boldsymbol{\alpha})}})))$ \label{alg:line:opt} \Comment{Optimize the linear bounds}
    \State $\polytope \gets \{ \mathbf{x} \mid \mathbf{x} \in \subdomain \wedge \lA^{(\boldsymbol{\alpha})} \mathbf{x} + \lb^{(\boldsymbol{\alpha})} \geq 0 \}$
    \State \Return $\polytope$ \Comment{Preimage approximation}
\EndFunction
~
\Function{EstimateApproxRatio}{$\polytopeset, f_\outputdomain$}
    \State $v_\polytope \gets \sum_{(\polytope,\subdomain,\samples) \in \polytopeset}\sum_{\xv \in \samples}\left[\xv \in \polytope\right] / |\samples| \cdot |\subdomain| $ \label{alg:line:volapp} \Comment{Approximation volume}
    \State $v_\outputdomain \gets \sum_{(\polytope,\subdomain,\samples) \in \polytopeset}\sum_{\xv \in \samples}\left[f_\outputdomain(\xv) \geq 0\right] / |\samples| \cdot |\subdomain| $ \label{alg:line:volpre} \Comment{Preimage volume}
    \State \Return $v_\polytope / v_\outputdomain$ \label{alg:line:approx}
\EndFunction
~
\Function{Priority}{$\polytope, \subdomain, \samples, f_\outputdomain$}
    \State $v_\polytope \gets \sum_{(\polytope,\subdomain,\samples) \in \polytopeset}\sum_{\xv \in \samples}\left[\xv \in \polytope\right] / |\samples| \cdot |\subdomain| $ \label{alg:line:volapp2}
    \State $v_\outputdomain \gets \sum_{(\polytope,\subdomain,\samples) \in \polytopeset}\sum_{\xv \in \samples}\left[f_\outputdomain(\xv) \geq 0\right] / |\samples| \cdot |\subdomain| $ \label{alg:line:volpre2}
    \State \Return $|v_\polytope - v_\outputdomain|$ \label{alg:line:volpri}

\EndFunction
~
\Function{SplitNeuron}{$\subdomain, \samples, l, i$}
    \State $\subdomain^-, \samples^- \gets \{\xv \mid \xv \in \subdomain \wedge \zv^l_i < 0\}, \{\xv \mid \xv \in \samples \wedge \zv^l_i < 0\}$ \label{alg:line:splitm}
    \State $\subdomain^+, \samples^+ \gets \{\xv \mid \xv \in \subdomain \wedge \zv^l_i \ge 0\}, \{\xv \mid \xv \in \samples \wedge \zv^l_i \ge 0\}$ \label{alg:line:splitp}
    \State \Return $\subdomain^-,\samples^-, \subdomain^+,\samples^+$ \Comment{Split domain and samples}
\EndFunction
\end{algorithmic}
\floatname{algorithm}{\scriptsize\textbf{Algorithm}}
\caption{\scriptsize
    Additional functions for \textsc{Premap2}, see \cref{alg:main} for details.
    The \lirpa\ function \citep{xu2020automatic} produces linear bounds with optimizable parameters in $\boldsymbol{\alpha}$ and
    $\zv^l_i$ is the activation of neuron~$i$ on layer~$l$.
}\label{alg:functions}
\end{algorithm}

\section{Efficiency and functionality improvements} \label{sec:methods}

In this section we provide the details of improvements and extensions implemented in PREMAP2. 
In \cref{sec:tighten,sec:sampling,sec:heuristics} we discuss algorithmic improvements over PREMAP, in \cref{sec:ci,sec:prior} new capabilities added to PREMAP2, and in \cref{sec:short,sec:batch} we describe computational strategies for increased efficiency.

\subsection{Bound tightening} \label{sec:tighten}

When a neuron is split, the linear relaxation error from that neuron to all the following layers is reduced.
However, \betacrown \citep{wang2021betacrown}, on which PREMAP is built, typically only updates the final layer because the optimization is computationally intensive \citep{wang2021betacrown}.
In PREMAP2, we propagate the bounds for intermediate layers using simpler LiRPA \citep{xu2020automatic} instead of using full \betacrown optimization.
While LiRPA only propagates bounds to the following layers in the network, a split also implies restrictions in previous layers.
Hence, we derive how the bounds of previous layers \emph{and the input} might tighten after a split, which helps future neuron selections and the sampling in \cref{sec:sampling}.

After splitting the $i$:th neuron on layer $l$, we use LiRPA \citep{xu2020automatic} as in \cref{eq:lirpa} to calculate the linear approximation bounds with respect to the preceding layers $m<l$ (including the input) such that 
\begin{equation*}
    \lA^{lm} \mathbf{z}^m + \lb^{lm} \leq \mathbf{z}^l \leq \uA^{lm} \mathbf{z}^m + \ub^{lm}.
\end{equation*}
The two constraints defined by the split, $\mathbf{z}^l_i<0$ and $\mathbf{z}^l_i\geq0$, can be propagated backwards using the linear approximations:
\begin{equation} \label{eq:split_approx}
[\lA^{lm} \mathbf{z}^m + \lb^{lm}]_i < 0 \textrm{~~and~~} 0 \leq [\uA^{lm} \mathbf{z}^m + \ub^{lm}]_i.
\end{equation}
From this we can update the bounds $\lz^m$ and $\uz^m$ so that:
\begin{alignat*}{2}
\lz^m_j \leftarrow& \max(\lz^m_j, -(c_j+\lb^{lm}_i)/\lA^{lm}_{ij}) &~\mid~ \lA^{lm}_{ij} < 0\\
\uz^m_j \leftarrow& \min(\uz^m_j, -(c_j+\lb^{lm}_i)/\lA^{lm}_{ij}) &~\mid~ \lA^{lm}_{ij} > 0
\end{alignat*}
for $\mathbf{z}^l_i < 0,$ where $c_j =\sum\nolimits_{k \neq j} \min(\lA^{lm}_{ik} \lz^m_k, \lA^{lm}_{ik}\uz^m_k)$ and
\begin{alignat*}{2}
\lz^m_j \leftarrow& \max(\lz^m_j, -(c_j+\ub^{lm}_i)/\uA^{lm}_{ij}) &~\mid~ \uA^{lm}_{ij} > 0 \\
\uz^m_j \leftarrow& \min(\uz^m_j, -(c_j+\ub^{lm}_i)/\uA^{lm}_{ij}) &~\mid~ \uA^{lm}_{ij} < 0
\end{alignat*}
for $\mathbf{z}^l_i \geq 0,$ where $c_j =\sum\nolimits_{k \neq j} \max(\uA^{lm}_{ik} \lz^m_k, \uA^{lm}_{ik}\uz^m_k)$.
After the ``reverse'' propagation of bounds we use LiRPA to update the bounds of subsequent layers.
This is done in \cref{alg:main} on \cref{alg:line:bounds}.
The effectiveness of the tighter bounds is evaluated in \cref{sec:ablation}.

\subsection{Heuristics} \label{sec:heuristics}

Judicious selection of an appropriate neuron to split (\cref{alg:main}, \cref{alg:line:select}) is crucial for the computational performance of preimage approximation.
This is achieved using 
a heuristic, which gives 
a score for each neuron and the neuron with the highest score is selected.
The original PREMAP \citep{zhang2024premap} aims to keep the splits balanced in terms of samples with a score function that counts the number of samples on each side of the split:
\begin{equation*}
    1 - \left|2\sum\nolimits_{i=1}^n[\mathbf{z}^l_{ij} \geq 0]/n - 1\right|,
\end{equation*}
where $\mathbf{z}_{ij}^l$ is the pre-activation for sample $j$ (out of $n$) at neuron $i$ on layer $l$.
However, this heuristic offers no indication of how well LiRPA \citep{xu2020automatic} and  \betacrown \citep{wang2021betacrown} might tighten the bounds.
Hence, for PREMAP2 we consider a range of alternative heuristics to calculate the priority score, detailed in \cref{tab:heur:desc}, and discuss how to select and combine them in \cref{sec:exp:params}.

\begin{table}[htb]
    \centering
    \caption{Heuristics for selecting which neuron to split}
    \label{tab:heur:desc}
    \begin{tabular}{r l l}
        \toprule
        \bfseries{Heuristic} & \bfseries{Formula} & \bfseries{From} \\
        \midrule
        balance & $1 - |2\sum_j[\zv^l_{ij} \geq 0]/n - 1|$ & \citet{zhang2024premap} \\
        soft & $1 - |2\sum_j\sigma(\zv^l_{ij})/n-1|$\\
        lower & $\max(-\lz^l_i, 0)$ & \citet{bunel2018unified} \\
        width & $\uz^l_i - \lz^l_i$ \\
        loose & $\uz^l_i - \lz^l_i - (\max_j(\zv^l_{ij}) - \min_j(\zv^l_{ij}))$ \\
        bound & $1 - (\max_j(\zv^l_{ij})-\min_j(\zv^l_{ij})) / (\uz^l_i - \lz^l_i)$ \\
        gap & $(-\lz^l_{i}\cdot\uz^l_{i}) / (\uz^l_i - \lz^l_i) $ & \citet{palma2021improved} \\
        area & $\sum_k |\lA^{lk}\lz^l_i\lz^l_i|$ \\
        under & $\sum_k |\lA^{lk}\lz^l_i|$ \\
        extra & $\sum_k\sum_{j} |\lA^{lk} \min(\zv^l_{ij}, 0)| / \sum_j[\zv^l_{ij} < 0]$ \\
        \bottomrule
    \end{tabular}
    \footnotetext{
        Unnormalized heuristics are, additionally, divided by their maximum value.
        The neuron (layer $l$, index $i$), with $j \in [1,\ldots,n]$ samples in $\samples$ and $k\in[1,\ldots,K]$ outputs of $f_\outputdomain$, with the highest score selected for splitting.
    }
\end{table}

The \emph{balance} \citep[from PREMAP,][]{zhang2024premap,zhang2024provable} and \emph{soft} heuristics prioritize neurons that would evenly split the samples.
The \emph{lower} \citep[used in \betacrown, \citealt{wang2021betacrown}]{bunel2018unified} and \emph{width} heuristics use the bounds given by LiRPA, while \emph{loose} and \emph{bound} compare these bounds with the minimum and maximum values achieved by the samples. 
Finally, the \emph{gap} \citep{palma2021improved},
\emph{area}, \emph{under}, and \emph{extra} heuristics measure how loose the linear relaxation is. 

Some heuristics in \cref{tab:heur:desc} use the (linear) bounds produced by LiRPA \citep{xu2020automatic,wang2021betacrown}.
To minimize computational overhead we cache them after each optimization.
Other heuristics use the activation values of the samples, which requires a forward pass through the network.
Therefore, no heuristic significantly extends the overall runtime.

We discuss 
the selection of which heuristics to use and their weights \cref{sec:exp:params}, but briefly remark here about their relative impact. Unsurprisingly, the heuristics related to how much the bounds can be tightened perform the best.
The \emph{gap} is visualized in the rightmost plot of \cref{fig:relu} as the vertical separation at zero.
For \emph{area}, \emph{under}, and \emph{extra} we use the linear bound with respect to the output of the whole (non-linear) network, as visualized in \cref{fig:heuristics}.
As shown in the figure, we only measure the ``relaxation error''  on negative values $\zv^l_j <= 0.0$ since those are, by definition, truncated to zero by the ReLU activation, while positive errors are non-linear.

\begin{figure}[htb]
    \centering
    \begin{tikzpicture}[every node/.style={scale=0.9}]
    \tikzset{scale=2.8}
        \filldraw[white!0, fill=oxyellow] (0.2,0) -- (1,0) -- (0.2,-0.35) -- (0.2,0);
        \draw[black!60, ->] (0,0) -- (1.6,0);
        \draw[black!60, <-] (1,0.5) -- (1,-0.4);
        \draw[very thick] (0.2,0) -- (1,0) -- (1.1, 0.3) -- (1.2, 0.2) -- (1.3, 0.4) -- (1.4, 0.2) -- (1.5,0.5);
        \draw (1.6, 0.1) node {$\zv^l_i$};
        \draw (0.75, 0.45) node {$f_\outputdomain(\zv^l_i)^k$};
        \draw[very thick, oxred] (0.2,-0.35) -- (1.5,0.22);
        \draw[dashed, oxred] (0.2,-0.35) -- (0.2,0.27);
        \draw[dashed, oxred] (1.5,-0.21) -- (1.5,0.5);
        \draw[oxred] (0.23, 0.38) node {$\lz^l_i$};
        \draw[oxred] (1.53, -0.3) node {$\uz^l_i$};
        \draw[oxred] (1.2, -0.1) node {$\lA^{lk}_i\zv^l_i$};
        \draw[ultra thick, oxgreen, <->] (0.15,-0.35) -- (0.15,0);
        \draw[ultra thick, oxaqua, {Circle[sep=-2pt,length=4pt]}-{Circle[sep=-2pt,length=4pt]}] (0.37,-0.27) -- (0.37,0) ;
        \draw[ultra thick, oxaqua, {Circle[sep=-2pt,length=4pt]}-{Circle[sep=-2pt,length=4pt]}] (0.5,-0.22) -- (0.5,0) ;
        \draw[ultra thick, oxaqua, {Circle[sep=-2pt,length=4pt]}-{Circle[sep=-2pt,length=4pt]}] (0.67,-0.15) -- (0.67,0) ;
    \end{tikzpicture}
    \caption{Visualizing three neuron selection heuristics: \emph{area} ({\color{oxyellow!80!black}yellow} shaded area), \emph{under} ({\color{oxgreen!80!black}green} distance), and \emph{extra} (the average {\color{oxaqua!80!black}blue} length at the samples).}
    \label{fig:heuristics}
\end{figure}

\subsection{Sampling} \label{sec:sampling}

PREMAP relies on sampling for Monte Carlo estimation of the volumes (\cref{alg:functions}, \crefrange{alg:line:volapp}{alg:line:volpri}) and the optimization of the bounds of \betacrown (on \cref{alg:line:opt}).
However, the original PREMAP \citep{zhang2024premap} only draws a large, fixed set of initial samples.
As each split subdivides the samples (\cref{alg:main}, \cref{alg:line:split}), the sizes of the subsets decrease exponentially (hence the use of the \emph{balance} heuristic).
Since a small number of samples means less reliable volume estimates and a greater chance of overfitting, 
in PREMAP2 we draw additional samples (\cref{alg:line:sample})
when the subsets get too small.

At first, we use rejection sampling from the input space,
but as the hit rate also decreases exponentially, we turn to \emph{hit-and-run} sampling \citep{montiel2013approximating,corte2021novel}.
The idea of \emph{hit-and-run} sampling is to start from a point inside a polytope, here based on the inequalities introduced in \cref{eq:split_approx},
and then sample a random direction, calculate the distance to the edge of the polytope, and uniformly sample a new point on this line.
This yields uniformly distributed samples \citep{montiel2013approximating} from the polytope and the procedure is GPU accelerated \citep{corte2021novel}.
Since consecutive steps are correlated, we take multiple steps between every recorded sample.
Note that \cref{eq:split_approx} is an outer bound, so we still invoke rejection sampling on the activations.

\subsection{Confidence intervals} \label{sec:ci}

As the volumes are calculated based on empirical distributions, they are not exact.
A more principled approach is to capture the uncertainty of volume estimation by calculating, e.g., confidence intervals. We add this new capability to PREMAP2.

When a neuron is split (on \cref{alg:line:split} of \cref{alg:main}) the volume $v$ of the new subdomains are proportional to $v_{+}\propto|\samples^+|/|\samples|$ and $v_{-}\propto|\samples^-|/|\samples|$ (\cref{alg:line:splitp,alg:line:splitm} of \cref{alg:functions}).
And for a branch with multiple splits: 
\begin{equation*}
    v_{*} =|\inputdomain| \cdot \frac{|\samples^*_1|}{|\samples_1|} \cdot \ldots \cdot \frac{|\samples^*_u|}{|\samples_u|},
\end{equation*}
where $|\inputdomain|$ is the total volume of the input domain. 
Hence, to define a distribution over $v_{*}$ we need to replace these fractions with distributions.

Since the volumes are based on samples, a practical approach is to use \emph{bootstrapping} \citep{diciccio1996bootstrap}.
To calculate a bootstrap distribution we resample the set of samples, with replacement, and recalculate the metrics (here volumes) multiple times (e.g., $1000$ times).
With the bootstrap distribution we can calculate
confidence intervals for individual splits and subdomains.

However, branches stemming from the same split are not independent.
So, for the preimage approximation as a whole ($v_\polytope$ in \cref{alg:functions}), we need additional considerations.
At every split, we must use the same bootstrap samples for both sides of the split.
Additionally, we use the same number of bootstrap samples for every split, so that we can combine the bootstrap values element-wise \emph{before} aggregation.
This preserves the dependencies not only for individual splits but also for deeper branches.

The precision of these volume estimates for the preimage $v_\outputdomain$ and approximation $v_\polytope$ increases with the number of samples $n$ (\cref{alg:main}, \cref{alg:line:sample}).
Therefore, we can use the confidence intervals to check if we have enough samples for a given precision.
This is empirically evaluated in \cref{sec:exp:ci} with encouraging results.

\subsection{Priors} \label{sec:prior}

The sampling procedure described \cref{sec:sampling} follows a uniform distribution.
This is also the case for related methods such as \citet{zhang2024premap,kotha2023provably}.
However, by giving each sample a non-negative weight we can define non-uniform input spaces.
Weights are also useful for applications where we need to prioritize, for example, certification of specific dangerous conditions. We thus enhance PREMAP2 with a capability of incorporating non-uniform prior distributions defined by the user.

To add \emph{optional} weights to PREMAP2, we replace the averages, $\sum\nolimits_{\xv\in\samples}\dottedbox/|\samples|$ on \cref{alg:line:opt,alg:line:volapp,alg:line:volpre,alg:line:volapp2,alg:line:volpre2} of \cref{alg:functions}, with weighted averages:
\begin{equation*}
    \sum_{\xv\in\samples}\dottedbox \cdot w(\xv)/\sum_{\xv\in\samples}w(\xv),
\end{equation*}
where $w \in \mathbb{R}^d\rightarrow\mathbb{R}_{\geq0}$ is a function producing a weight for any input.
This adds a small overhead compared to plain sums, but does not affect the computational complexity.

The weight function $w$ need not be a normalized distribution. However, if the weight is highly concentrated in a small area or changes rapidly, we may require many samples for an accurate representation.
A common diagnostic method for weighted empirical distributions is to calculate the \emph{effective sample size} \citep{kish1992weighting}:
\begin{equation} \label{eq:ess}
    n_\textrm{eff} = \frac{\left(\sum_{\xv\in\samples} w(\xv)\right)^2}{\sum_{\xv\in\samples} w(\xv)^2}.
\end{equation}
If $n_\textrm{eff} \ll n$ (say $n_\textrm{eff} < 0.01n$) then the implicit distribution defined by $w$ deviates substantially from the uniform distribution,
and we would need a very large number of samples, $n$, to obtain reliable volume estimates.
We calculate $n_\textrm{eff}$ on the use cases in \cref{sec:exp:weights} and discuss the results and implications for weighted PREMAP2.

\subsection{Shortcuts} \label{sec:short}

In some cases, a branch can be discarded or finalized before running the heavy \betacrown\ process, for example, when the branch contains no preimage samples (when under-approximating) or only preimage samples (when over-approximating).
This enables computational optimization by avoiding potentially heavy computation (in \cref{alg:main}, on \cref{alg:line:left} or \ref{alg:line:right}) on that branch.
For the remaining branch we can perform an additional split before running \betacrown.

Furthermore, if all samples are assigned to one branch of a split it means that either (a) the other branch is impossible to reach, but the current LiRPA bounds are not tight enough, or (b) the other side is reachable, but our current estimate for that volume is near zero.
We can detect these splits with just a single forward pass through the network, which lets us split and stabilize all of them simultaneously on \cref{alg:line:shortcut} of \cref{alg:main} (before invoking \betacrown).

We implement these shortcuts in PREMAP2 to optimize computational performance, evaluated in \cref{sec:ablation}.

\subsection{Batching} \label{sec:batch}

The original PREMAP \citep{zhang2024premap,zhang2024provable} generates approximations of a split in parallel (see \cref{alg:main}, \cref{alg:line:left,alg:line:right}).
In addition, \betacrown \citep{wang2021betacrown} processes the branches (the loop on \cref{alg:line:loop}) in batches to take advantage of GPU resources.
The pseudocode shows neither for readability, but in practice PREMAP2 uses both strategies.

On easy problems (that require few splits) batching offers no significant advantage.
However, if finding tighter bounds for the selected branch is difficult (\cref{alg:line:pop,alg:line:select} of \cref{alg:line:loop}), processing secondary branches in the same batch can lead to faster progress and earlier stopping (due to the criteria on \cref{alg:line:loop}).
On hard problems (that require many splits), the secondary branches have to be explored anyway and the purpose of batching is to maximize the use of available compute resources.
Batching is then limited by the (linear) increase in memory usage.

\section{Empirical evaluation} \label{sec:exp}

In this section we empirically evaluate the quality of our improvements
implemented in PREMAP2.
In \cref{sec:exp:setup} we describe the experimental setup and datasets, whereas in \cref{sec:exp:params} the selection of the weights for the new heuristics.
We compare PREMAP2 to the original PREMAP on patch attacks and reinforcement learning tasks in \cref{sec:exp:comp} and on different network architectures in \cref{sec:exp:net}.
Finally, in \cref{sec:ablation} we perform an ablation study.

The code for both PREMAP2 and the experiments is available under an open source licence\footnote{Available at: \url{https://github.com/Anton-Bjorklund/Premap2}}.

\subsection{Experimental setup} \label{sec:exp:setup}

We ran all experiments (including the use cases in \cref{sec:usecases})
with four cores from an Intel Xeon 6252 CPU and one NVIDIA RTX 2080 Ti.
All preimage approximations were limited to 10 minutes with a batch size of 2.
Unless otherwise stated,
we use a stopping threshold for the approximation ratio of $0.9$ for under-approximations and $1.1$ for over-approximations.

We certify robustness against patch attacks on the following image datasets: GTSRB \citep[traffic signs,][]{stallkamp2012gtsrb}, SVHN \citep[house numbers,][]{netzer2011svhn}, and CIFAR-10 \citep[general images,][]{krizhevsky2009cifar}.
Details about the model architectures and training (a fully connected network with three $300$-wide layers and a convolutional network with three convolutional layers and one fully connected layer) are included in Appendix~\ref{app:networks}.
For experiments with image datasets ($32\times32$ pixels) we randomly sample images, evenly from every class.
The patches are rectangular, with sizes between $3 \times 3$ and $6 \times 6$, and are randomly placed within the images.

In \cref{sec:exp:comp}, we use the same five \citep{julian2019reachability,muller2023third,brockman2016openai} (low-dimensional) reinforcement learning control tasks and networks from \citet{zhang2024premap}.
We compute preimage approximations for selected actions (classes), such as \emph{fire main engine} in lunarlander and \emph{push left} in cartpole, for the same input domains as in \citet{zhang2024premap} (with ${x_v\in[-2,0]}$ for lunarlander, ${x_\omega\in[-2,0]}$ for cartpole, and ${x_v\in[-0.1,0.1]}$ for dubinsrejoin).
Every setup is repeated ten times with different seeds.

When running the original PREMAP, we use a batch size of 1 (as it does not support batching). 
We also backported the necessary changes from PREMAP2 to enable PREMAP on CNNs, by making PREMAP agnostic to the additional dimensions in the input and activation tensors and making use of the support for convolutions in LiRPA \citep{xu2020automatic} and \betacrown \citep{wang2021betacrown}.

\subsection{Hyperparameters} \label{sec:exp:params}

In this section we discuss how to select and combine the heuristics discussed in \cref{sec:heuristics}.
We use the tasks described in \cref{sec:exp:setup}. 
For each task we uniformly sample multiple weights $\boldsymbol{w}$ for the heuristics so that half are inactive and at least one has a weight of one:
\begin{align*}
\boldsymbol{w}' &\sim \mathbb{U}[-1, 1]\\
\boldsymbol{w} &= \max(\boldsymbol{0}, \boldsymbol{w}') / \max(\boldsymbol{w}'),
\end{align*}
and perform both over- and under-approximations.

To evaluate the heuristics, we calculate the difference in the approximation ratio (\cref{alg:functions}, \cref{alg:line:approx}) from the first optimization $v_{\polytope 1}/{v_\outputdomain}$ (\cref{alg:main}, \cref{alg:line:preimage}) to the final iteration $v_{\polytope \textrm{f}}/{v_\outputdomain}$ and divide by the total time $t_\textrm{f}$ (with a 5 minute time limit):
\begin{equation}
   \label{eq:cov_delta}
   \delta = \left|\frac{v_{\polytope 1}}{v_\outputdomain}-\frac{v_{\polytope \textrm{f}}}{v_\outputdomain}\right|/t_\textrm{f}.
\end{equation}
Higher values of $\delta$ indicate that the approximation ratio improves more quickly.
To compare across tasks and datasets, we normalize the $\delta$ values by dividing by the maximum $\delta$ for each task.
We also filter out tasks where the heuristic choice makes no difference, such as when PREMAP2 finishes after the first optimization ($\textrm{f}=1$).

\begin{figure}[htb]
    \centering
    \includegraphics[scale=0.6]{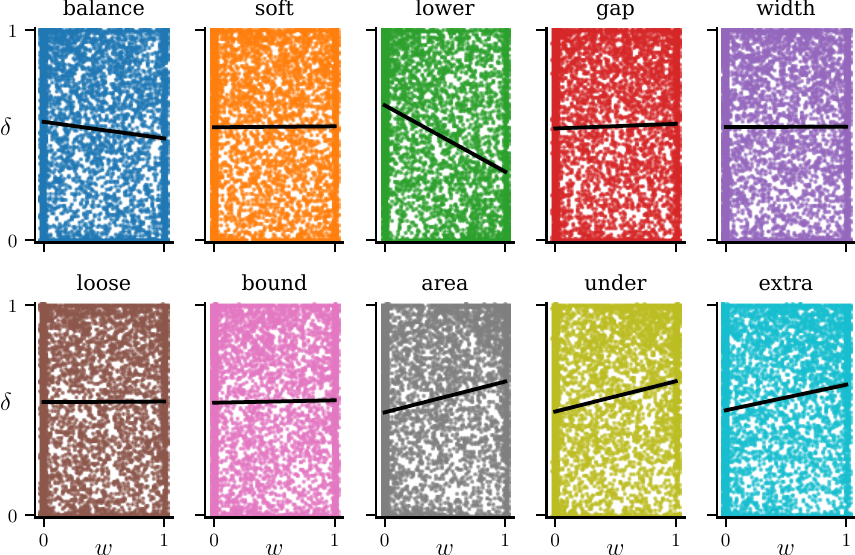}
    \caption{
    The effect of using various heuristics for the selection of which neuron to split (each dot represents a different combination).
    The black line is a single variable, linear regression,  where a positive slope is correlated with faster improvements, see \cref{eq:cov_delta}.
    }
    \label{fig:heuristic_dists}
\end{figure}

The results for individual heuristics can be seen in \cref{fig:heuristic_dists}, with a linear model showing the general trend.
Combinations with the \emph{area}, \emph{under}, and \emph{extra} heuristics enabled are correlated with faster improvements to the approximation ratio, while using \emph{balance} \citep{zhang2024premap} and \emph{lower} \citep{bunel2018unified} instead of other heuristics leads to slower approximations.

However, some heuristics are correlated and some might complement each other, so we cannot choose the combination based on individual plots.
To evaluate how heuristics combine we fit a kernel ridge regression ensemble, using a grid search with cross-validation to set the parameters ($\alpha=8$, $\gamma=1.0$, and a Laplacian kernel).
In \cref{tab:weights} we evaluate the model on a grid, $\boldsymbol{w} \in \{0.0, 0.25, 0.5, 0.75, 1.0\}^{10}$, and show the best results for increasing numbers of non-zero coefficients for the different heuristics.

\begin{table}[ht]
    \centering
    \caption{Comparing different combinations of heuristics}
    \label{tab:weights}
    \sisetup{detect-weight, mode=text}
    \begin{tabular}{@{~}S[table-format = 1.3(4)] r@{~}r@{~}r@{~}r@{~}r@{~}r@{~}r@{~}r@{~}r@{~}r@{~}}
        \toprule
        \multicolumn{1}{c}{$\boldsymbol\delta$} & \multicolumn{2}{c}{\bfseries{Weights}} \\
        \midrule
        0.804(0.010) &  extra: & $1.00$, &  area: & $0.75$ \\
        0.785(0.011) &  extra: & $1.00$, &  area: & $1.00$ \\
        0.783(0.008) &  under: & $1.00$, &   gap: & $0.75$ \\
        0.780(0.009) &  extra: & $1.00$, &  area: & $0.50$ \\
        \midrule
        0.832(0.010) &  extra: & $1.00$, &  area: & $0.75$, &   gap: & $0.25$ \\
        0.827(0.008) &  extra: & $1.00$, &  area: & $0.75$, &   gap: & $0.50$ \\
        0.817(0.011) &  extra: & $1.00$, &  area: & $0.75$, & under: & $0.50$ \\
        0.817(0.009) &  under: & $1.00$, &  area: & $0.75$, &  soft: & $0.50$ \\
        \midrule
        \bm 0.837(0.010) &  extra: & $1.00$, &  area: & $0.75$, & under: & $0.50$, &   gap: & $0.25$ \\
        0.836(0.010) &  extra: & $1.00$, &  area: & $0.75$, & under: & $0.25$, &   gap: & $0.25$ \\
        0.833(0.009) &  extra: & $1.00$, & under: & $0.75$, &  area: & $0.75$, &   gap: & $0.50$ \\
        0.832(0.009) &  extra: & $1.00$, & under: & $0.75$, &  area: & $0.75$, &   gap: & $0.25$ \\
        \midrule
        0.831(0.010) &  extra: & $1.00$, &  area: & $0.75$, & under: & $0.25$, &   gap: & $0.25$, &  soft: & $0.25$ \\
        0.831(0.007) &   soft: & $1.00$, & extra: & $0.75$, &  area: & $0.75$, & under: & $0.50$, &   gap: & $0.25$ \\
        0.830(0.009) &  extra: & $1.00$, &  area: & $0.75$, & under: & $0.50$, &   gap: & $0.25$, &  soft: & $0.25$ \\
        0.829(0.006) &   soft: & $1.00$, & extra: & $0.75$, & under: & $0.75$, &  area: & $0.75$, &   gap: & $0.25$ \\
        \bottomrule
    \end{tabular}
    \footnotetext{We evaluate combinations heuristics by training ten kernel ridge regression models (using k-fold splits) to predict $\delta$ (larger is better). Showing the best combinations for increasing number of non-zero weights.}
\end{table}

Using too few heuristics is suboptimal, but enabling too many dilutes the signal.
As shown in \cref{tab:weights}, the sweet spot is between three and five. 
For the rest of the experiments in this paper we use the following heuristics and weights: \emph{extra} at $1.0$, \emph{area} at $0.75$, \emph{under} at $0.5$, and \emph{gap} at $0.25$.
Separate results for under- and over-approximations can be seen in Appendix~\ref{app:heuristics}.

\subsection{Confidence intervals} \label{sec:exp:ci}

\begin{figure}[htb]
    \centering
    \includegraphics[scale=0.6]{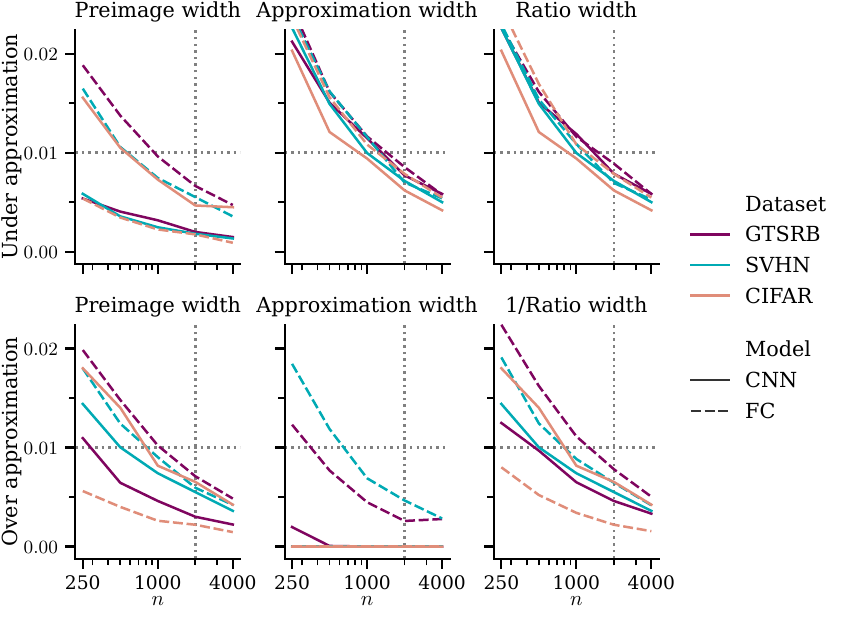}
    \caption{
    Widths of the confidence intervals as the number of samples increases.
    The plotted curves represent the $90$th percentiles for the preimage volume $v_\outputdomain$, approximation volume $v_\polytope$, and the approximation ratio.
    Over-approximations are shows as the inverse ratio to keep the unit between one and zero.
    }
    \label{fig:ci}
\end{figure}

We next investigate the impact of the choice of the number of samples $n$ (\cref{alg:main}, \cref{alg:line:sample}) using the confidence intervals discussed in \cref{sec:ci}.
In \cref{fig:ci} we plot the width of the confidence intervals by calculating the distance between the lower and upper bounds for random patches with increasing number of samples.
As the approximation ratios for over-approximations are unbounded above one, we show the inverse ratio so that all estimates map to the same scale between zero and one.
The plotted curves in \cref{fig:ci} represent the $90$th percentile of confidence intervals, 
meaning that for most tasks we obtain intervals less than one percent wide with $n=2000,$ which we use as the default for all the experiments in this paper. 

\begin{figure}[htb]
    \centering
    \includegraphics[scale=0.6]{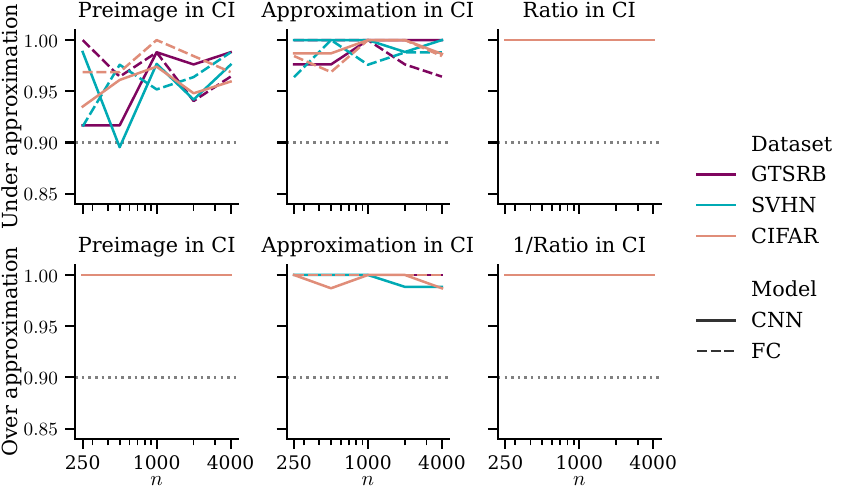}
    \caption{Fraction of volume and ratio estimates (non-bootstrap) that fall within the 90 \% bootstrap confidence intervals as the number of samples increases.}
    \label{fig:ci2}
\end{figure}

In \cref{fig:ci2} we verify that the non-bootstrap volume estimates fall within the bootstrap confidence intervals.
Since we use 90\% confidence intervals for both figures, we expect the intervals to hold at least that often, which is what we observe in \cref{fig:ci2}.

\subsection{Comparison} \label{sec:exp:comp}

In this section we compare PREMAP2 with the original PREMAP \citep{zhang2024premap}.
In \cref{tab:comparison} we consider the five reinforcement learning tasks from \citet{zhang2024premap}.
However, as this paper is focused on ReLU splitting, we explicitly run the tasks with ReLU splitting, even if input splitting might be better suited for some tasks \citep[\emph{cartpole} and \emph{dubinsrejoin}]{zhang2024premap}. 
We run these tasks purely on the CPU, since a GPU introduces significant overhead due to the small size of the models.
Each task is run ten times with different seeds for both under- and over-approximations.

\begin{table}[htb]
    \centering
    \caption{Comparing PREMAP2 on tabular data}
    \label{tab:comparison}
    \sisetup{detect-weight, mode=text}
    \addtolength{\tabcolsep}{-0.1em}
    \begin{tabular}{l@{~~}l S[table-format = 1.3(4)]@{~~}c@{~}lS[table-format = 4.1(4)]S[table-format = 3.2(4)]}
        \toprule
        \bfseries{Method} & \bfseries{Dataset} & \multicolumn{3}{c}{\hspace{-5mm}\bfseries{Approximation ratio}} & \bfseries{Iterations} & \bfseries{Time (s)} \\
        \midrule
        PREMAP  & auto\_park   & \bm 0.915(0.006) &     $\ge$ & \bm  0.9 &     10.5( 0.8) &     0.94(0.09) \\
                & vcas         & \bm 0.905(0.003) &     $\ge$ & \bm  0.9 &     31.1( 3.1) &     3.85(0.22) \\
                & lunarlander  &     0.488(0.009) & $\not\ge$ &     0.75 &     4930.2(126.6) &     600.21(0.13) \\
                & cartpole     &     0.266(0.011) & $\not\ge$ &     0.75 &     4661.5(108.3) &     600.21(0.13) \\
                & dubinsrejoin &     0.000(0.000) & $\not\ge$ &     0.75 &     3725.1(79.5) &     600.25(0.12) \\
        PREMAP2 & auto\_park   & \bm 1.000(0.000) &     $\ge$ & \bm  0.9 & \bm  3.9( 0.3) & \bm 0.21(0.01) \\
                & vcas         & \bm 0.964(0.002) &     $\ge$ & \bm  0.9 & \bm  7.0( 0.0) & \bm 0.80(0.04) \\
                & lunarlander  & \bm 0.750(0.001) &     $\ge$ & \bm 0.75 & \bm 171.8( 8.8) & \bm 29.97(1.71) \\
                & cartpole     & \bm 0.776(0.014) &     $\ge$ & \bm 0.75 & \bm 15.7( 0.5) & \bm 20.55(1.24) \\
                & dubinsrejoin & \bm 0.465(0.008) &     $\ge$ & \bm 0.75 & \bm 1421.4(48.1) &     595.97(1.26) \\
        \midrule
        \multicolumn{7}{c}{Under-approximation $\uparrow$ \hspace{2em}
          \begin{tikzpicture}[overlay]\draw[line cap=round] (-.16,-.16)--(.32,.32);\end{tikzpicture}
          \hspace{2em} $\downarrow$ Over-approximation} \\
        \midrule
        PREMAP  & auto\_park   & \bm 1.032(0.005) &     $\le$ & \bm  1.1 &     14.0( 0.0) &     2.06(0.13) \\
                & vcas         & \bm 1.059(0.002) &     $\le$ & \bm  1.1 & \bm  1.0( 0.0) &     0.08(0.01) \\
                & lunarlander  & \bm 1.232(0.009) &     $\le$ & \bm 1.25 &     53.4( 5.1) &     8.40(0.77) \\
                & cartpole     & \bm 1.237(0.007) &     $\le$ & \bm 1.25 &     98.3( 5.9) &     11.47(0.90) \\
                & dubinsrejoin & \bm 1.250(0.000) &     $\le$ & \bm 1.25 &     1865.6(171.3) &     221.08(20.28) \\
        PREMAP2 & auto\_park   & \bm 1.016(0.002) &     $\le$ & \bm  1.1 & \bm  3.0( 0.0) & \bm 0.20(0.01) \\
                & vcas         & \bm 1.058(0.002) &     $\le$ & \bm  1.1 & \bm  1.0( 0.0) & \bm 0.06(0.00) \\
                & lunarlander  & \bm 1.195(0.022) &     $\le$ & \bm 1.25 & \bm 10.5( 0.7) & \bm 1.55(0.10) \\
                & cartpole     & \bm 1.235(0.006) &     $\le$ & \bm 1.25 & \bm 10.2( 0.4) & \bm 3.08(0.59) \\
                & dubinsrejoin & \bm 1.238(0.016) &     $\le$ & \bm 1.25 & \bm 10.1( 2.1) & \bm 2.12(0.41) \\
        \bottomrule
    \end{tabular}
    \footnotetext{
    Comparing PREMAP2 to PREMAP on five reinforcement learning controllers. 
    The time limit is ten minutes (or when the target threshold is reached) and the results are averaged from ten different seeds.
    We measure the approximation ratio (larger/smaller is better for under/over approximations), 
    the number of subdomains required to reach that coverage (smaller is better), 
    and the time it took (smaller is better).
    }
\end{table}

Over-approximations are easier for both versions, and PREMAP2 is consistently faster for both under- and over-approximations.
PREMAP2 only exceeds the 10-minute time limit when under-approximating \textit{dubinsrejoin}, but still makes more progress (higher approximation ratio).

\begin{table}[htb]
    \centering
    \caption{Comparing PREMAP2, PREMAP and ProbSpec on patch attacks}
    \label{tab:comp_gtsrb}
    \sisetup{detect-weight, mode=text}
    \addtolength{\tabcolsep}{-0.1em}
    \begin{tabular}{@{}l@{}c@{~}S[table-format = 1.2(1.2)]@{~}S[table-format = 2.2(2.2)]S[table-format = 1.2(1.2)]@{~}S[table-format = 1.2(1.2)]S[table-format = 3.1(3.1)]@{~}S[table-format = 3.1(3.1)]@{}}
        \toprule
        \bfseries{Method} & \bfseries{Model}  & \multicolumn{2}{@{}c@{}}{\bfseries{Approx. ratio}} & \multicolumn{2}{@{}c@{}}{\bfseries{Completed}} & \multicolumn{2}{@{}c@{}}{\bfseries{Time (s)}} \\
       \multicolumn{2}{c}{Dataset: GTSRB} & {All} & {Hard} & {All} & {Hard} & {All} & {Hard} \\
        \midrule
        PREMAP   & FC  &     0.54 \pm 0.48 &     0.37 \pm 0.45 &     0.50 \pm 0.50 &     0.32\pm 0.47 &     309.9 \pm 295.3 &     419.9 \pm 269.6 \\
                 & CNN &     0.37 \pm 0.48 &     0.31 \pm 0.46 &     0.36 \pm 0.48 &     0.30\pm 0.46 &     392.1 \pm 279.6 &     425.1 \pm 267.1 \\
        PREMAP2  & FC  & \bm 0.88 \pm 0.25 & \bm 0.84 \pm 0.27 & \bm 0.85 \pm 0.36 & \bm 0.79\pm 0.41 & \bm 108.5 \pm 213.6 & \bm 146.2 \pm 237.6 \\
                 & CNN & \bm 0.71 \pm 0.41 & \bm 0.69 \pm 0.42 & \bm 0.70 \pm 0.46 & \bm 0.68\pm 0.47 & \bm 203.8 \pm 266.9 & \bm 221.1 \pm 271.3 \\
        ProbSpec & FC  &     0.31 \pm 0.46 &     0.05 \pm 0.21 &     0.29 \pm 0.46 &     0.04\pm 0.19 &     396.1 \pm 251.1 &     539.7 \pm  89.8 \\
                 & CNN &     0.12 \pm 0.33 &     0.05 \pm 0.20 &     0.13 \pm 0.33 &     0.05\pm 0.21 &     487.7 \pm 178.0 &     531.0 \pm 106.6 \\
        \midrule
        \multicolumn{8}{c}{Under-approximation $\uparrow$ \hspace{2em}
          \begin{tikzpicture}[overlay]\draw[line cap=round] (-.16,-.16)--(.32,.32);\end{tikzpicture}
          \hspace{2em} $\downarrow$ Over-approximation} \\
        \midrule
        PREMAP   & FC  &     1.16 \pm 0.63 &      2.05 \pm  1.34 &     0.86 \pm 0.35 &     0.04 \pm 0.21 &     89.3 \pm 204.2 &     576.4 \pm 22.5 \\
                 & CNN &     1.87 \pm 3.84 &      8.65 \pm  9.05 &     0.89 \pm 0.32 &     0.00 \pm 0.00 &     75.1 \pm 172.2 &     555.4 \pm 10.9 \\
        PREMAP2  & FC  & \bm 1.05 \pm 0.26 & \bm  1.30 \pm  0.62 & \bm 0.94 \pm 0.24 & \bm 0.61 \pm 0.50 & \bm 41.7 \pm 144.8 & \bm 268.9 \pm 288.6 \\
                 & CNN & \bm 1.12 \pm 0.56 & \bm  1.94 \pm  1.38 & \bm 0.94 \pm 0.24 & \bm 0.50 \pm 0.51 & \bm 39.0 \pm 141.5 & \bm 298.1 \pm 303.6 \\
        ProbSpec & FC  &     2.07 \pm 7.17 &      8.18 \pm 17.71 &     0.85 \pm 0.36 &     0.00 \pm 0.00 &     82.4 \pm 196.8 &     552.3 \pm 10.7 \\
                 & CNN &     2.79 \pm 9.17 &     17.61 \pm 23.72 &     0.89 \pm 0.31 &     0.00 \pm 0.00 &     61.4 \pm 175.5 &     564.4 \pm 28.6 \\
        \bottomrule
    \end{tabular}
    \footnotetext{
    Comparing PREMAP2, PREMAP, and ProbSpec on the GTSRB dataset.
    The ``Hard'' columns exclude tasks that reach the threshold in the first iteration.
    We measure the average approximation ratio (larger/smaller is better for under/over approximations), the number of instances reaching the threshold within 10 minutes (larger is better), and the average time required (smaller is better).
    }
\end{table}

\cref{tab:comp_gtsrb} shows the results for the GTSRB dataset on both a fully connected (FC) and a convolutional neural network (CNN).
We also include a comparison to \textit{ProbSpec} \citep{boetius2025solving}, which is another method for preimage approximations.
We use the same, approximation-ratio-based, stopping criterion as for PREMAP2 and limit the batch size to the capacity of the GPU.
However, as ProbSpec only uses input splitting \citep{boetius2025solving} rather than ReLU splitting, we do not expect it to scale well to large input spaces and models.

Some preimage approximations are easy and finish without a single split (\cref{alg:main}, \cref{alg:line:loop}).
Therefore, in \cref{tab:comp_gtsrb} we show separately the results for all cases (columns labelled ``All'') and cases where at least one of the methods required splitting (columns labelled ``Hard'').
We observe that 
PREMAP2 is able to complete almost twice the number of under-approximations compared to PREMAP (within the 10-minute time limit), and offers a greater advantage on the ``Hard'' cases.
Under-approximations are often more computationally challenging than over-approximations.
As expected, due to the use of input splitting, ProbSpec \citep{boetius2025solving} struggles with under-approximations for any case that requires splitting (including almost all ``Hard'' cases).

\subsection{Network architectures} \label{sec:exp:net}

\begin{figure}[htb]
    \centering
    \includegraphics[scale=0.6]{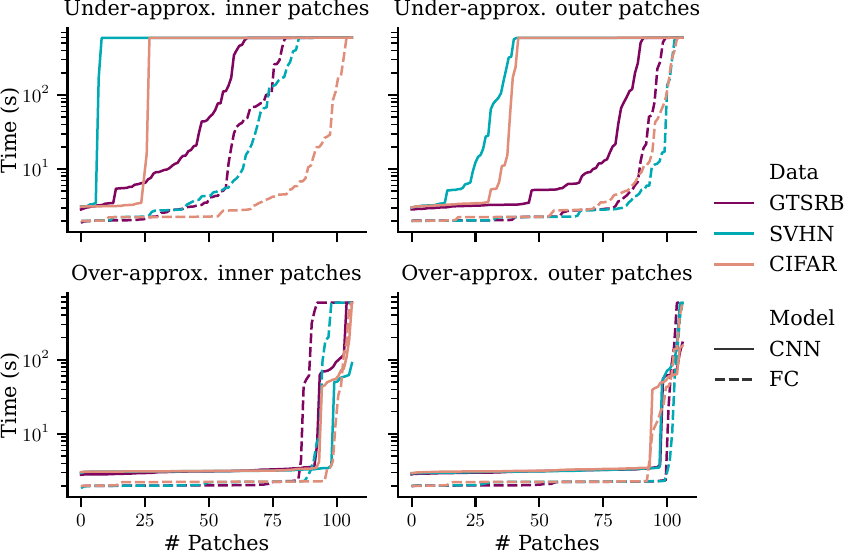}
    \caption{Time to calculate a preimage under-approximation with a 10-minute time limit and $0.9$ volume threshold. The patches are applied to random images (from every class) with random positions and sizes. We group the results based on whether the centre of the patch is within 10 pixels of the centre (inner vs outer patch).}
    \label{fig:comparison}
\end{figure}

Next we evaluate PREMAP2 on different neural network architectures. 
In \cref{fig:comparison} we show plots for PREMAP2 both fully connected and convolutional neural networks trained on image datasets.
We divide the results into two groups, based on whether the patch is close to the edge of the image (outer patch) or positioned centrally (inner patch).

For under-approximations, both GTSRB and SVHN show strong dependence on the location of the patch, which is to be expected since the most influential parts of the image (traffic signs and digits) are in the centre of the images.
Some SVHN images even have distracting digits in the surroundings \citep{netzer2011svhn}, which further emphasizes the centre during training.

The LiRPA processing time and the maximum number of splits is dependent on the number of unstable neurons.
For CNNs, the number of activations scales with the size of the image and is substantial:
for example, just the first layer has $30\times30\times32 = 28~800$ activations. 
As can be seen in \cref{fig:comparison}, this makes the CNNs slower and more challenging than fully connected networks, even though they have fewer parameters.

Since most non-adversarial perturbations do not change the predicted class,
the preimages tend to be quite large (close to 100\% coverage).
This makes many over-approximations trivial as the initial domain already has approximation ratio below $1.1$, especially for the more robust CNNs.

\subsection{Ablation study} \label{sec:ablation}

\begin{figure}[htb]
    \centering
    \includegraphics[scale=0.6]{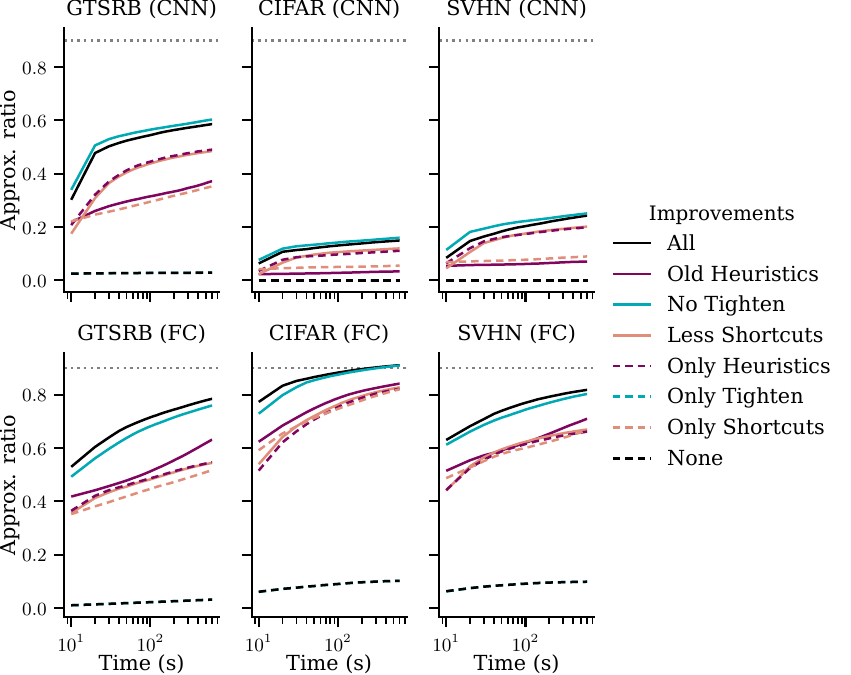}
    \caption{Comparing the average under-approximation ratio (higher is better) over time, with different improvements enabled or disabled.
    }
    \label{fig:ablation_under}
\end{figure}

\begin{figure}[htb]
    \centering
    \includegraphics[scale=0.6]{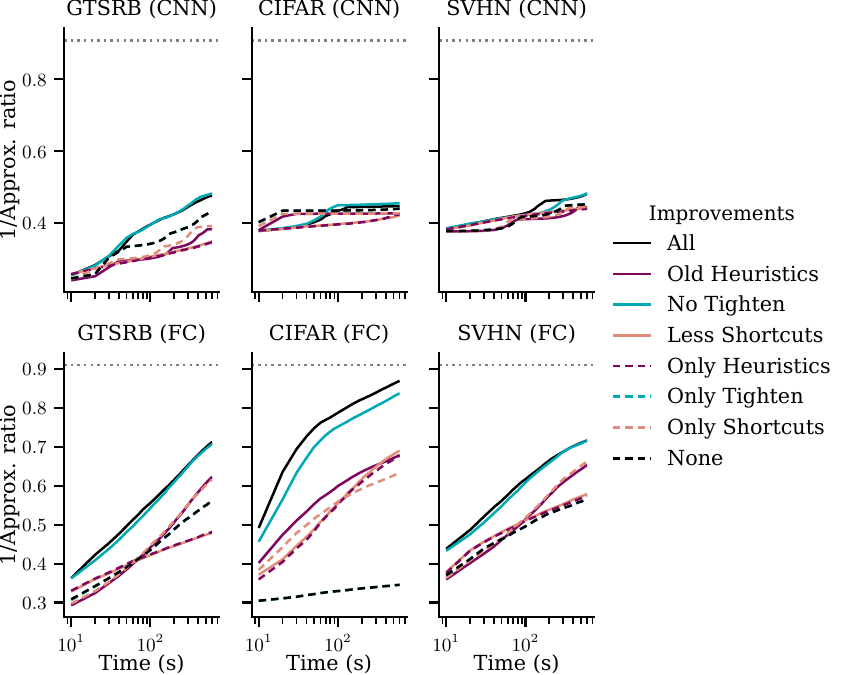}
    \caption{Comparing the average inverse over-approximation ratio (higher is better) over time, with different improvements enabled or disabled.
    }
    \label{fig:ablation_over}
\end{figure}

In this section we perform an ablation study on (a) the tightening of intermediate bounds, (b) the new split selection heuristics, and (c) the pre-emptive shortcuts, respectively discussed in \cref{sec:tighten,sec:heuristics,sec:short}.
The results can be seen in \cref{fig:ablation_under,fig:ablation_over}, where we measure the approximation ratio over time for the three image datasets with both types of networks and random patches, as before.
We average the results after removing any task that finishes after the first optimization (since the improvements do not matter in that case).

For the under-approximations in \cref{fig:ablation_under}, the new heuristics offer the best improvements, but the shortcuts are also very helpful.
Tightening of intermediate layer bounds has a minor effect, since some of the benefit is countered by the additional computational overhead, especially for CNNs.
Hence, for larger networks (where LiRPA slows down), it might be beneficial to disable this improvement.

For the over-approximations in \cref{fig:ablation_over}, we observe similar trends regarding the new heuristics and shortcuts providing the best improvements. 
The over-approximation results also contained more instances
that finish after one iteration.

\section{Use cases} \label{sec:usecases}

In this section we demonstrate a range of use cases for PREMAP2, showing how to exploit over- and under-approximations when analysing the performance of neural networks.
And in \cref{sec:patch}, we extend the discussion of the motivating example of certification against patch attacks by demonstrating how to use PREMAP2 with a wider range of patches.
In \cref{sec:ood} we extend the outlier detection use case from \citet{kotha2023provably} and show how, with PREMAP2, we are able to certify more properties than previously by using both under- and over-approximations.
Then, in \cref{sec:xai} we investigate a
novel use case in explainable artificial intelligence (XAI), where
we use PREMAP2 to highlight pixels that are important for the correct classification of a given image.
Finally, in \cref{sec:exp:weights} we demonstrate how non-uniform priors can be used to certify fairness.

For the experiments on the use cases, we utilize
the same synthetic dataset and network as in \citet{kotha2023provably} for outlier detection; create a new dataset from GTSRB \citep{stallkamp2012gtsrb} and CIFAR-10 \citep{krizhevsky2009cifar} for evaluating explanations; and utilize the same model and prior as in \citet{boetius2025solving} for the fairness example.

\subsection{Patch attack robustness} \label{sec:patch}

\begin{figure}[htb]
    \centering
    \includegraphics[scale=0.59]{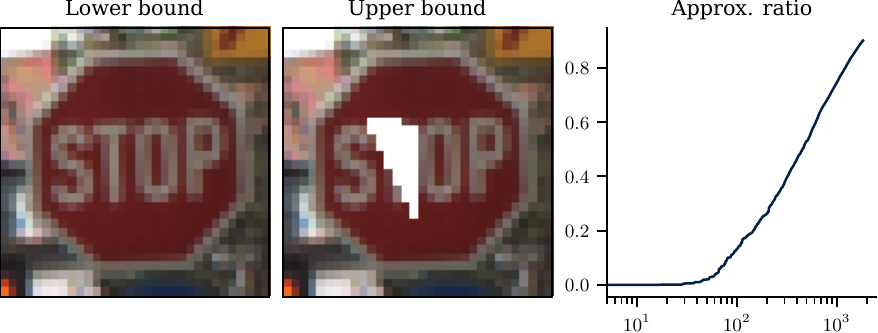}
    \caption{Certifying robustness against a non-rectangular patch. The first two images show the lower and upper bounds of the input domain, $\inputdomain$, and the plot on the right shows progress of PREMAP2 over time in log scale.
    }
    \label{fig:lightbeam}
\end{figure}

Robustness certification of patch attacks is intractable for existing methods due to the scaling required.
As demonstrated in \cref{sec:exp:net}, PREMAP2 is able to handle rectangular patches on both fully connected and convolutional networks. 
However, PREMAP2 is flexible enough to handle a wider range of patch shapes. 
\cref{fig:lightbeam} demonstrates a patch inspired by the sunbeam in \cref{fig:ex1} (middle).
We use a stop sign from the GTSRB \citep{stallkamp2012gtsrb} dataset.
Since a sunbeam only adds brightness, we define the lower bound of the input domain, $\inputdomain$, to be the selected image and the upper bound to contain a patch with maximum possible brightness (white) of an irregular shape (pixelated triangle).

\cref{fig:lightbeam} (right) also demonstrates the anytime property of the PREMAP algorithm:
stopping early still ensures an under-approximation, just a smaller one, and over time more approximations approach the $0.9$ volume threshold.
The curve is also characteristic of PREMAP2.
There is an initial delay before the bounds become tight enough to certify any coverage and the improvements slow down as the subdomains reduce in size (note the log scale).

\subsection{Out-of-distribution detection} \label{sec:ood}

In \citet{kotha2023provably} the authors propose that preimage over-approximations as computed by tool INVPROP can be used for certified out-of-distribution (OOD) detection.
During training an additional logit is added to represent the out-of-distribution data points that are added to the training data \citep{hendrycks2019deep}.
Comparing that logit with the logits for the real classes allows for detection of outliers.

The over-approximations of INVPROP \citep{kotha2023provably} divide the input-space into \emph{certified outliers} and \emph{probable non-outliers}.
With PREMAP2 we are able to add a new category: \emph{certified non-outliers}.
This is due to PREMAP2 being able to compute both under- and over-approximations.

\begin{figure}[htb]
    \centering
    \includegraphics[scale=0.6]{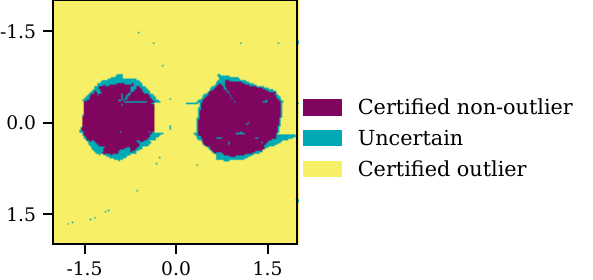}
    \caption{Using preimages for certifying out-of-distribution detection.}
    \label{fig:ood}
\end{figure}

In \cref{fig:ood} we use the same model as in \citet{kotha2023provably}, a fully connected neural network with layer sizes $200$, $200$, and $3$.
The model was trained on non-outliers from two, spatially separated, normal distributions.
The outliers were generated by uniformly sampling from the input space, $[-2,2]^2$, and removing any sample too close to a non-outlier.
With PREMAP2 we obtain a $0.9$ under-approximation of the non-outliers in $6.9$ seconds and a $1.1$ over-approximation in $2.9$ seconds when running only on the CPU.
For reference, in \citet{kotha2023provably} INVPROP achieved an over-approximation of $1.14$ in $12.0$ seconds utilizing a GPU.

\subsection{Explainability (XAI)} \label{sec:xai}

XAI aims to provide analysis tools to investigate predictions, e.g., 
to aid in model development and debugging \citep{barredoarrieta2020explainable}.
A popular approach in XAI is to cover parts of an image and observe if the classification changes \citep{ivanovs2021perturbationbased}.
The reasoning is easy to understand: if the prediction changes that area must have been important.
However, this requires careful design of the replacement pattern \citep{fong2017interpretable}.
By using PREMAP2 we do not have to consider a singular replacement pattern at a time, but rather all possible replacements.

\begin{figure}[htb]
    \centering
    \includegraphics[scale=0.6]{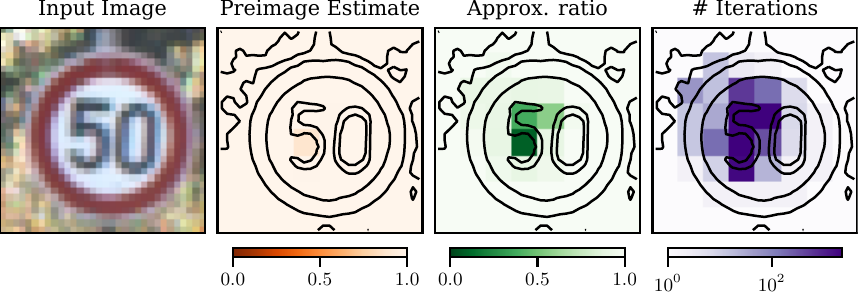}
    \caption{Using PREMAP2 to investigate which parts of the image are important for classification.
    The figures are, from left to right: the original image, the estimated preimage volume (higher means more robust), the approximation ratio of the preimage, and the number of iterations (lower means easier).
    }
    \label{fig:xai}
\end{figure}

In \cref{fig:xai} we apply $4\times4$ patches on a grid covering the image on the left.
In the second image from the left we show the estimated volume of the preimage (not the approximation) for every patch location.
By using PREMAP2, we can quantify the proportion of all possible patches that would change the prediction, which allows us to rank the importance of different parts of the image.
In this case, we see a patch near the `5' being slightly highlighted.

Furthermore, in the two images on the right of \cref{fig:xai}, we plot the approximation ratio and the number of iterations before the 10-minute time limit, respectively.
More areas around the digit `5' are highlighted, indicating that approximating the preimage is more difficult in this region of the image.
This difficulty is due to 
the model being highly non-linear in those areas \citep{black2022interpreting}
(because of other speed limit signs that only differ in the first digit).

\begin{figure}[htb]
    \centering
    \includegraphics[scale=0.6]{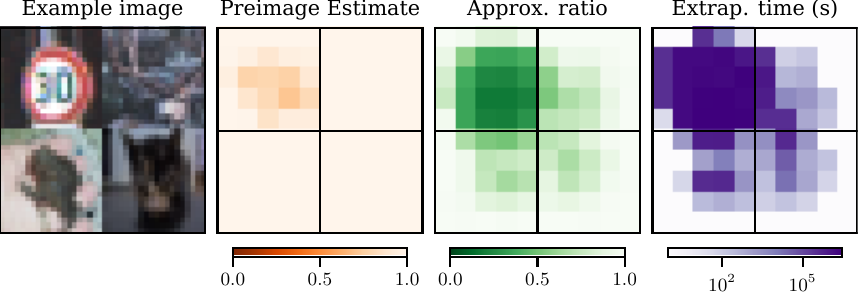}
    \caption{
    Aggregating multiple explanations on a dataset combining four images.
    The figures are (from left to right):  example image, the average preimage volume, the average approximation ratio, and the (extrapolated) time required to reach an approximation ratio of $0.9$.
    }
    \label{fig:xai_grid}
\end{figure}

To quantitatively evaluate explanations, we follow \citet{anani2025pixellevel} and create a dataset where we combine one labelled image (from GTSRB) with three unrelated images (from CIFAR-10) and train a CNN classifier.
The leftmost image of \cref{fig:xai_grid} shows an example.
This allows us to control the location of the meaningful part of the image: a random quadrant during training and top left for evaluation.

In the second image of \cref{fig:xai_grid} we visualize the mean preimage volume from multiple explanations.
All the highlighted patches are in the correct quadrant.
However, in roughly one fifth of all explanations we find no patch able to change the prediction.
In these cases we might want a secondary indicator of importance, such as the time required by PREMAP2.
When the time limit is exceeded, we use the achieved approximation ratio to extrapolate the time required to reach the threshold of $0.9$ (shown in the rightmost image).

\begin{figure}[htb]
    \centering
    \includegraphics[scale=0.6]{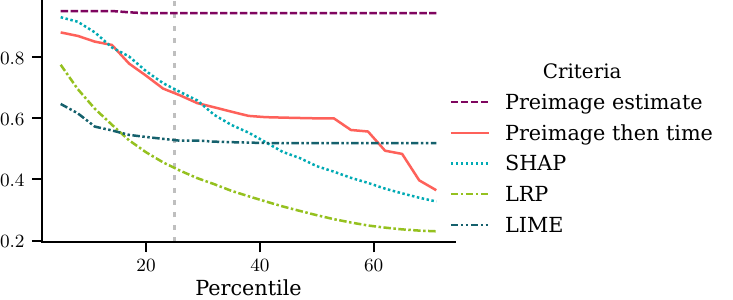}
    \caption{%
    Counting how many of the most important pixels, according to different explanations, are in the correct quadrant for 20 different images.
    Higher is better below 25\% (the size of one quadrant).
    }
    \label{fig:xai_comp}
\end{figure}

In \cref{fig:xai_comp} we compare the explanations from PREMAP2 to three dedicated XAI methods on 20 new images (generated from randomly selected GTSRB images with different classes). 
We use the partition version of SHAP \citep{lundberg2017unified} with $1000$ samples, which produces patches of similar size to what we use for PREMAP2, and 
``inpaint\_telea'' as the replacement pattern.
For LIME \citep{ribeiro2016why}, we specify manual ``superpixels'' that match the grid we used for PREMAP2 and the default grey replacement.
Finally, we use LRP \citep{bach2015pixelwise,montavon2018methods} that produces pixel-level importance values.
For each explanation method we rank the pixels by importance and 
count the fraction of pixels that are within the correct quadrant for increasing percentiles.

The pixels highlighted using the preimage volumes are always in the correct quadrant, but some explanations lack any rankings.
When we use time as a secondary criterion,
the results are close to the best XAI method, SHAP.
The LRP explanations are highly granular, often focusing on edges, including edges outside the correct quadrant.
LIME also highlights a lot of irrelevant pixels.
The dedicated XAI methods are faster, taking at most a couple of seconds. 
However, the preimage-based ranking not only yields a competitive importance score, but also quantifies how sensitive the prediction is in different parts of the image.

\subsection{Weighted samples} \label{sec:exp:weights}

As mention in  \cref{sec:prior}, we can use weights both to prioritize important inputs and to implement priors for PREMAP2.
In this section we demonstrate both use cases.

\begin{figure}[htb]
    \centering
    \includegraphics[scale=0.6]{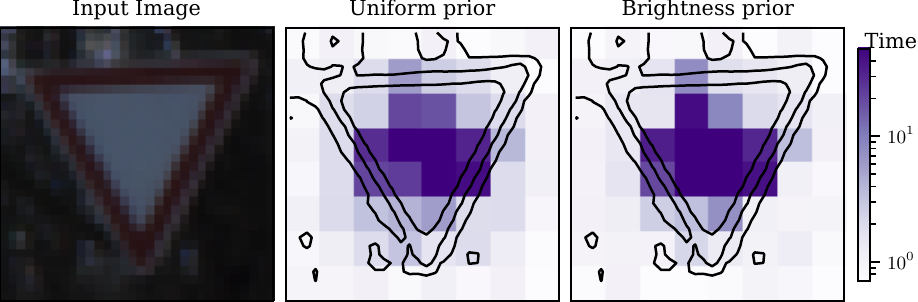}
    \caption{Comparing PREMAP2 results with (right) and without (middle) weights that prioritize pixels with the same brightness as the original image (left).
    We measure the time required to reach the approximation threshold (with the same extrapolation as in \cref{fig:xai_grid}). }
    \label{fig:prior}
\end{figure}

First, we describe how PREMAP2 can be used to prioritize patches with realistic pixels.
The picture in \cref{fig:prior} (left) depicts a dark image.
A real-world patch, captured at the same time by the same camera, would likely have the same limited brightness.
Hence, for certifying defences against patch attacks, we could limit the brightness in the input domain, $\inputdomain$.
However, this would not account for all adversarial patches, e.g., patches incorporating reflectors or lights.
In \cref{fig:prior} (right) we allow pixels to take all colours between pure white and pure black, but prioritize pixels conforming to the overall brightness of the image with weights given by
\begin{equation*}
    \prod_{(x,y)\in\text{Patch}} 1-\frac{ \max(0, \overline{p}_{xy}-\overline{p}_{\max}) }{ 1-\overline{p}_{\max} },
\end{equation*}
where $\overline{p}_{xy}$ is the brightness (average colour) of a pixel and $\overline{p}_{max}$ the brightness of the brightest pixel.
All patches have an effective samples size (ESS), see \cref{eq:ess}, of at least 0.3, which means that the weighted distribution is well supported by the uniform samples. 

Compared to the uniform prior modelled with no weights (\cref{fig:prior} middle), the weight-prioritized brightness prior results place even less emphasis on patches outside the traffic sign.
This implies that grater changes are needed in these regions to have any effect on the model prediction, which is in line with previous findings in \cref{fig:comparison}.

Our second use case, summarized in 
\cref{tab:prior}, replicates a fairness experiment from \citet{boetius2025solving}, using the same model and prior as that defined in \citet{albarghouthi2017fairsquare}. 
We consider the most challenging setup with a Bayesian network prior and a neural network using three variables and two hidden neurons.
Since this is a small model, we also train an additional network based on five variables with two layers of ten and five hidden neurons.
The target is to predict high income and we follow \citet{boetius2025solving} by investigating the demographic parity for males versus females, when the models are trained without that variable.

To run PREMAP2 and ProbSpec, we also need to define lower and upper limits for the variables.
The limits used in \citet{boetius2025solving} are very loose and result in very small ESS, \cref{eq:ess}, due to the prior having slim tails for extreme values, which map poorly to uniform samples.
Instead, we use limits that are closer to the limits of the dataset (for both methods), which yields small, but acceptable, ESS.

\begin{table}[h]
    \centering
    \caption{Preimage approximations with a prior}
    \label{tab:prior}
    \sisetup{detect-weight, mode=text}
    \begin{tabular}{l@{~~}l@{~~}l S[table-format=3.2(4)]@{~~}S[table-format=1.3(4)]@{~~}S[table-format=1.3(4)]@{~~}S[table-format=1.3(4)]}
        \toprule
        \bfseries{Model} & \bfseries{Method} & \bfseries{Condition}\hspace{-2mm} & \bfseries{Time} & \bfseries{Under} & \bfseries{Over} & \bfseries{ESS} \\
        \midrule
        $\text{NN}_{3,2}$    & PREMAP2  & Female & \bm   0.75( 0.40) & \bm 0.584(0.026) & \bm 0.584(0.026) & 0.038(0.002) \\
                             &          & Male   & \bm   0.41( 0.03) & \bm 0.695(0.015) & \bm 0.695(0.015) & 0.069(0.003) \\
                             & ProbSpec & Female &       1.18( 0.11) &     0.564(0.000) &     0.574(0.000) & \\
                             &          & Male   &       1.21( 0.11) &     0.678(0.000) &     0.688(0.000) & \\
        $\text{NN}_{5,10,5}$ & PREMAP2  & Female & \bm  35.52(23.92) & \bm 0.511(0.015) & \bm 0.532(0.029) & 0.038(0.002) \\
                             &          & Male   & \bm  32.67(23.02) & \bm 0.669(0.012) & \bm 0.685(0.013) & 0.069(0.003) \\
                             & ProbSpec & Female &     600.20( 0.11) &     0.464(0.000) &     0.554(0.000) & \\
                             &          & Male   &     600.21( 0.09) &     0.633(0.000) &     0.708(0.000) & \\
        \bottomrule
    \end{tabular}
    \footnotetext{Using PREMAP2 and ProbSpec for checking group fairness (demographic parity) on two small networks trained without the sensitive variable and a Bayesian network as the prior. Lower times and tighter bounds is better.}
\end{table}

For the smaller network, the PREMAP2 results in \cref{tab:prior} replicates \citet{boetius2025solving}, where males have a slight advantage, but the difference is below the threshold ($0.15$).
We use a stopping criterion of $0.99$, which leads to PREMAP2 splitting both neurons, decomposing the neural network into exact linear segments.
This results in the upper bound matching the lower bound.
However, running the experiment multiple times with different seeds results in slight variations in the Monte Carlo volume estimates, which is unsurprising given the small ESS.
ProbSpec \citep{boetius2025solving} uses axis-aligned input splits, which leaves small gaps between the lower and upper bounds.
ProbSpec provides both under- and over-approximations in the same execution, but also takes more than twice the time. 

On the larger network, PREMAP2 finishes in less than a minute, while ProbSpec exceeds the 10 minute time limit, leading to a larger gap between the lower and upper bound.
With PREMAP2 we can conclude that the model always violates the parity threshold (comparing the female over-approximation to the male under-approximation), while the ProbSpec result is inconclusive since the bounds are too loose.

\section{Conclusions} \label{sec:conc}

We have developed PREMAP2, which substantially improves the PREMAP framework \citep{zhang2024provable,zhang2024premap} for computing preimage approximations for neural networks through a collection of algorithmic extensions and a demonstration of its enhanced usability through a broad range of use cases.  

In \cref{sec:methods}, we carefully devised a set of 
heuristics for selecting regions to split, improved bound tightening, and implemented
efficiency improvements that systematically outperform the original PREMAP on 
both smaller benchmarks and patch attack certification on images, as well as ensuring tractability of certification for convolutional neural networks.
We also extended PREMAP with new capabilities, in the form of confidence intervals and support for non-uniform prior distributions. 

In \cref{sec:exp}, we evaluated the improvements and compared PREMAP2 to related methods, whereas, in \cref{sec:usecases}, we 
demonstrated how the ability to compute both under- and over-approximations can be utilized in certification of adversarial robustness, outlier detection, explainability, and fairness.

Through systematic experimental evaluation, we have demonstrated the superiority of PREMAP2 over its competitors and thus its readiness for deployment in neural network development, debugging and certification. 
Future prospects for PREMAP2 include adapting to more network architectures, such as transformers, graph neural networks, and larger convolutional networks.
On the applied side, it would be beneficial to extend certification to further domains, for example, biometric security or medical diagnosis.
Here, confidence intervals can serve as an important tool for increasing trust in model predictions and the addition of priors unlocks further applications.

\section*{Declarations}
\addcontentsline{toc}{section}{Declarations}

\begin{itemize}
    \item \textbf{Funding:}
This project received funding from the ERC under the European Union’s Horizon 2020 research and innovation programme (FUN2MODEL, grant agreement No.~834115) and ELSA: European Lighthouse on Secure and Safe AI project (grant agreement No.~101070617 under UK guarantee).

\item \textbf{Competing interests:}
The authors have no competing or financial interests to declare that are relevant to the content of this article. 

\item \textbf{Data availability:}
All datasets used in this article are openly available, see the cited sources for details. 

\item \textbf{Code availability:}
The code for both PREMAP2 and the experiments is available under an open source licence at: \url{https://github.com/Anton-Bjorklund/Premap2}.

\item \textbf{Author contribution:}
A.B. and M.K. wrote the main manuscript text.
A.B designed and implemented PREMAP2 and performed all experiments.
M.Z. performed a preliminary investigation into patch attacks during an internship at the University of Oxford.
P.M. co-designed the method and experiments for non-uniform priors.
M.K. also reviewed the design of PREMAP2.
All authors reviewed the manuscript.

\end{itemize}

\begin{appendices}

\section{Heuristics} \label{app:heuristics}

\cref{tab:weights_under,tab:weights_over} show the heuristic selection results similar to \cref{tab:weights} but with under- and over-approximations separated (using the same hyperparameters of $\alpha=8$, $\gamma=1.0$, and a Laplacian kernel).
Both tables show similar results to \cref{tab:weights} in \cref{sec:exp:params}, prioritizing the same heuristics.
The over-approximations show lower predicted $\delta$, which means that the choice of heuristics has a lesser impact on the results.
This is unsurprising considering the evidence in \cref{tab:comparison,fig:comparison} showing that over-approximations are easier.

\begin{table}[htb]
    \centering
    \caption{Combinations of heuristics for under-approximations}
    \label{tab:weights_under}
    \sisetup{detect-weight, mode=text}
    \begin{tabular}{@{}S[table-format = 1.3(4)] r@{~}r@{~}r@{~}r@{~}r@{~}r@{~}r@{~}r@{~}r@{~}r@{}}
        \toprule
        \multicolumn{1}{c}{$\boldsymbol\delta$} & \multicolumn{2}{c}{\bfseries{Weights}} \\
        \midrule
        0.863(0.012) &  extra: & $1.00$, &  area: & $1.00$ \\
        0.858(0.006) &   area: & $1.00$, &   gap: & $0.75$ \\
        0.854(0.008) &   area: & $1.00$, &   gap: & $0.50$ \\
        0.853(0.013) &  extra: & $1.00$, &  area: & $0.75$ \\
        \midrule
        \bm 0.890(0.009) &   area: & $1.00$, & extra: & $0.75$, &   gap: & $0.50$ \\
        0.887(0.010) &  extra: & $1.00$, &  area: & $1.00$, &   gap: & $0.50$ \\
        0.882(0.011) &  extra: & $1.00$, &  area: & $1.00$, &   gap: & $0.25$ \\
        0.878(0.011) &  under: & $1.00$, &  area: & $0.75$, &  soft: & $0.50$ \\
        \midrule
        0.886(0.009) &   area: & $1.00$, & extra: & $0.75$, &   gap: & $0.50$, & under: & $0.25$ \\
        0.885(0.007) &   area: & $1.00$, & extra: & $0.50$, & under: & $0.50$, &   gap: & $0.50$ \\
        0.884(0.009) &   area: & $1.00$, & under: & $0.75$, &   gap: & $0.50$, &  soft: & $0.50$ \\
        0.884(0.010) &  under: & $1.00$, &  area: & $0.75$, &   gap: & $0.50$, &  soft: & $0.50$ \\
        \midrule
        0.882(0.008) &  under: & $1.00$, &  area: & $0.75$, &   gap: & $0.50$, &  soft: & $0.50$, & extra: & $0.25$ \\
        0.882(0.007) &   area: & $1.00$, & under: & $0.75$, &   gap: & $0.50$, &  soft: & $0.50$, & extra: & $0.25$ \\
        0.882(0.006) &   area: & $1.00$, & under: & $0.75$, &   gap: & $0.50$, &  soft: & $0.50$, & width: & $0.25$ \\
        0.881(0.006) &   area: & $1.00$, & extra: & $0.50$, & under: & $0.50$, &   gap: & $0.50$, &  soft: & $0.25$ \\
        \bottomrule
    \end{tabular}
\end{table}

\begin{table}[htb]
    \centering
    \caption{Combinations of heuristics for over-approximations}
    \label{tab:weights_over}
    \sisetup{detect-weight, mode=text}
    \begin{tabular}{@{}S[table-format = 1.3(4)] r@{~}r@{~}r@{~}r@{~}r@{~}r@{~}r@{~}r@{~}r@{~}r@{}}
        \toprule
        \multicolumn{1}{c}{$\boldsymbol\delta$} & \multicolumn{2}{c}{\bfseries{Weights}} \\
        \midrule
        0.669(0.012) &  extra: & $1.00$, &  area: & $0.75$ \\
        0.647(0.009) &  extra: & $1.00$, &  area: & $1.00$ \\
        0.644(0.011) &  extra: & $1.00$, & under: & $1.00$ \\
        0.640(0.010) &  extra: & $1.00$, &  area: & $0.50$ \\
        \midrule
        0.688(0.012) &  extra: & $1.00$, &  area: & $0.75$, & under: & $0.50$ \\
        0.685(0.008) &  extra: & $1.00$, & under: & $0.75$, &  area: & $0.75$ \\
        0.682(0.005) &  under: & $1.00$, &  area: & $0.75$, & loose: & $0.75$ \\
        0.682(0.012) &  extra: & $1.00$, &  area: & $0.75$, &  soft: & $0.25$ \\
        \midrule
        \bm 0.699(0.009) &  extra: & $1.00$, & under: & $0.75$, &  area: & $0.75$, &   gap: & $0.25$ \\
        0.697(0.012) &  extra: & $1.00$, &  area: & $0.75$, & under: & $0.50$, &   gap: & $0.25$ \\
        0.693(0.008) &  extra: & $1.00$, & under: & $0.75$, &  area: & $0.50$, &   gap: & $0.25$ \\
        0.692(0.012) &  extra: & $1.00$, &  area: & $0.75$, & under: & $0.50$, &  soft: & $0.25$ \\
        \midrule
        0.691(0.008) &  extra: & $1.00$, & under: & $0.75$, &  area: & $0.75$, &   gap: & $0.25$, & balance: & $0.25$ \\
        0.690(0.010) &  extra: & $1.00$, & under: & $0.75$, &  area: & $0.75$, &   gap: & $0.25$, & loose: & $0.25$ \\
        0.690(0.011) &  extra: & $1.00$, &  area: & $0.75$, & under: & $0.50$, &   gap: & $0.25$, &  soft: & $0.25$ \\
        0.689(0.009) &  extra: & $1.00$, &  area: & $0.75$, & under: & $0.50$, &   gap: & $0.25$, & balance: & $0.25$ \\
        \bottomrule
    \end{tabular}
\end{table}

\section{Neural networks} \label{app:networks}

For the image datasets we use two types of neural networks: fully connected and convolutional.
The architectures are described in \cref{tab:arch}.
The networks were trained with cross-entropy loss using the AdamW \citep{loshchilov2019decoupled} optimizer
with brightness and hue variation until the accuracy on the test set stopped decreasing (usually around 50 epochs).
For the reinforcement learning tasks we used the same networks as in \citet{zhang2024premap}, for the out-of-distribution detection we used the same network as in \citet{kotha2023provably}, and for the fairness prior experiment we used the same network as in \citet{albarghouthi2017fairsquare} and a fully connected network with two hidden layers (of sizes 10 and 5).

\begin{table}[htb]
    \centering
    \caption{Architectures for the neural networks}
    \label{tab:arch}
    \begin{tabular}{lc c lc}
        \toprule
        \multicolumn{2}{c}{\bfseries{Fully connected}} && \multicolumn{2}{c}{\bfseries{Convolutional}} \\
        \bfseries{Layer} & \bfseries{Output size} && \bfseries{Layer} & \bfseries{Output size} \\
        \midrule
        Input & $32 \times 32 \times 3$ && Input & $32 \times 32 \times 3$ \\
        Flatten & $3072$ && Convolutional & $30 \times 30 \times 3$ \\
        Linear & $300$ && ReLU & $30 \times 30 \times 32$ \\
        ReLU & $300$ && Average pooling & $15 \times 15 \times 32$ \\
        Linear & $300$ && Convolutional & $12 \times 12 \times 32$ \\
        ReLU & $300$ && ReLU & $12 \times 12 \times 32$ \\
        Linear & $300$ && Average pooling & $6 \times 6 \times 32$ \\
        ReLU & $300$ && Convolutional & $4 \times 4 \times 64$ \\
        Dropout & $300$ && Average pooling & $2 \times 2 \times 64$ \\
        Linear & $|\mathbf{y}|$ && ReLU & $2 \times 2 \times 64$ \\
        &&& Flatten & $256$ \\
        &&& Dropout & $256$ \\
        &&& Linear & $|\mathbf{y}|$ \\
        \bottomrule
    \end{tabular}
\end{table}

\end{appendices}

\bibliography{references}

\end{document}